\documentclass[10pt,twocolumn,letterpaper]{article}

\usepackage{iccv}
\usepackage{times}
\usepackage{epsfig}
\usepackage{graphicx}
\usepackage{amsmath}
\usepackage{amssymb}
\usepackage{color}
\usepackage{xspace}
\usepackage{subfig}
\usepackage{float}
\usepackage{contour}
\usepackage{xkeyval}
\usepackage{comment}
\usepackage{tabularx}
\usepackage{authblk}

\usepackage[pagebackref=true,breaklinks=true,letterpaper=true,colorlinks,bookmarks=false]{hyperref}

\definecolor{darkgreen}{RGB}{0,127,0}
\definecolor{darkblue}{RGB}{0,0,127}
\definecolor{darkmagenta}{RGB}{127,0,127}
\definecolor{darkred}{RGB}{127,0,0}
\definecolor{darkcyan}{RGB}{0,127,127}
\definecolor{gray}{RGB}{100,100,100}

\iccvfinalcopy %

\def\iccvPaperID{759} %

\def\etal{\emph{et al.}\xspace}
\newcommand{\boldhead}[1]{\vspace{0.045in}\noindent\textbf{\frenchspacing{#1.}}\hspace{0.2cm}}

\begin{document}

\newcommand{\todo} [1] {\textcolor{darkred}{[TODO: #1]}} 
\newcommand{\note} [1] {\textcolor{darkblue}{[note: #1]}} 

\newcommand{\wh}{\color{white}} %
\contourlength{.5pt}
\newcommand{\figlabel}[1]{\sffamily\bfseries\wh\scriptsize\contour{black}{#1}} %
\makeatletter
\newlength{\sfp@hseplen}\newlength{\sfp@vseplen}
\define@cmdkey{subfigpos}[sfp@]{vsep}[0.6\baselineskip]{\setlength{\sfp@vseplen}{\sfp@vsep}}%
\define@cmdkey{subfigpos}[sfp@]{hsep}[2.5pt]{\setlength{\sfp@hseplen}{\sfp@hsep}}%
\newcommand{\subfigimg}[3][,]{%
  \setkeys{Gin,subfigpos}{vsep,hsep,#1}%
  \setbox1=\hbox{\includegraphics{#3}}%
  \leavevmode\rlap{\usebox1}%
  \rlap{\hspace*{8pt}\raisebox{\dimexpr\ht1-7pt}{\figlabel{#2}}}%
  \phantom{\usebox1}%
}
\makeatother

\title{3D Scene Reconstruction with Multi-layer Depth and Epipolar Transformers}

\author[1]{\vspace{-0.2cm}Daeyun Shin}
\author[2]{Zhile Ren}
\author[1]{Erik B. Sudderth}
\author[1]{Charless C. Fowlkes}
\affil[ ]{

\textsuperscript{1}University of California, Irvine
\hspace{0.9em}
\textsuperscript{2}Georgia Institute of Technology
\vspace{0.15cm}
}
\affil[ ]{\small{\url{https://research.dshin.org/iccv19/multi-layer-depth}}
\vspace{-0.5cm}
}
\setlength{\affilsep}{-1.1em}
\renewcommand\Authsep{\hspace{1.8em}}
\renewcommand\Authands{\hspace{1.8em}}

\maketitle

\begin{abstract}
We tackle the problem of automatically reconstructing a complete 3D model of a scene
from a single RGB image. This challenging task requires inferring the shape of
both visible and occluded surfaces.  Our approach utilizes
viewer-centered, multi-layer representation of scene geometry adapted from
recent methods for single object shape completion.  To improve the accuracy of
view-centered representations for complex scenes, we introduce a novel
``Epipolar Feature Transformer'' that transfers convolutional network features
from an input view to other virtual camera viewpoints, and thus better covers
the 3D scene geometry. Unlike existing approaches that first detect and localize
objects in 3D, and then infer object shape using category-specific models,
our approach is fully convolutional, end-to-end differentiable, and
avoids the resolution and memory limitations of voxel representations.  We
demonstrate the advantages of multi-layer depth representations and epipolar
feature transformers on the reconstruction of a large database of indoor
scenes.
\end{abstract}

\setlength{\textfloatsep}{10pt}
\section{Introduction}

When we examine a photograph of a scene, we not only perceive the 3D shape of
visible surfaces, but effortlessly infer the existence of many invisible surfaces.
We can make strong predictions about the complete shapes of familiar objects
despite viewing only a single, partially occluded aspect, and can infer
information about the overall volumetric occupancy with sufficient accuracy
to plan navigation and interactions with complex scenes.  This remains a
daunting visual task for machines despite much recent progress in detecting
individual objects and making predictions about their shape.  \emph{Convolutional
neural networks} (CNNs) have proven incredibly successful as tools for learning
rich representations of object identity which are invariant to intra-category
variations in appearance. Predicting 3D shape rather than object category has
proven more challenging since the output space is higher dimensional and
carries more structure than simple regression or classification tasks.

\begin{figure}[t]
\begin{center}
\vspace{-0.03in}
  \includegraphics[width=\columnwidth]{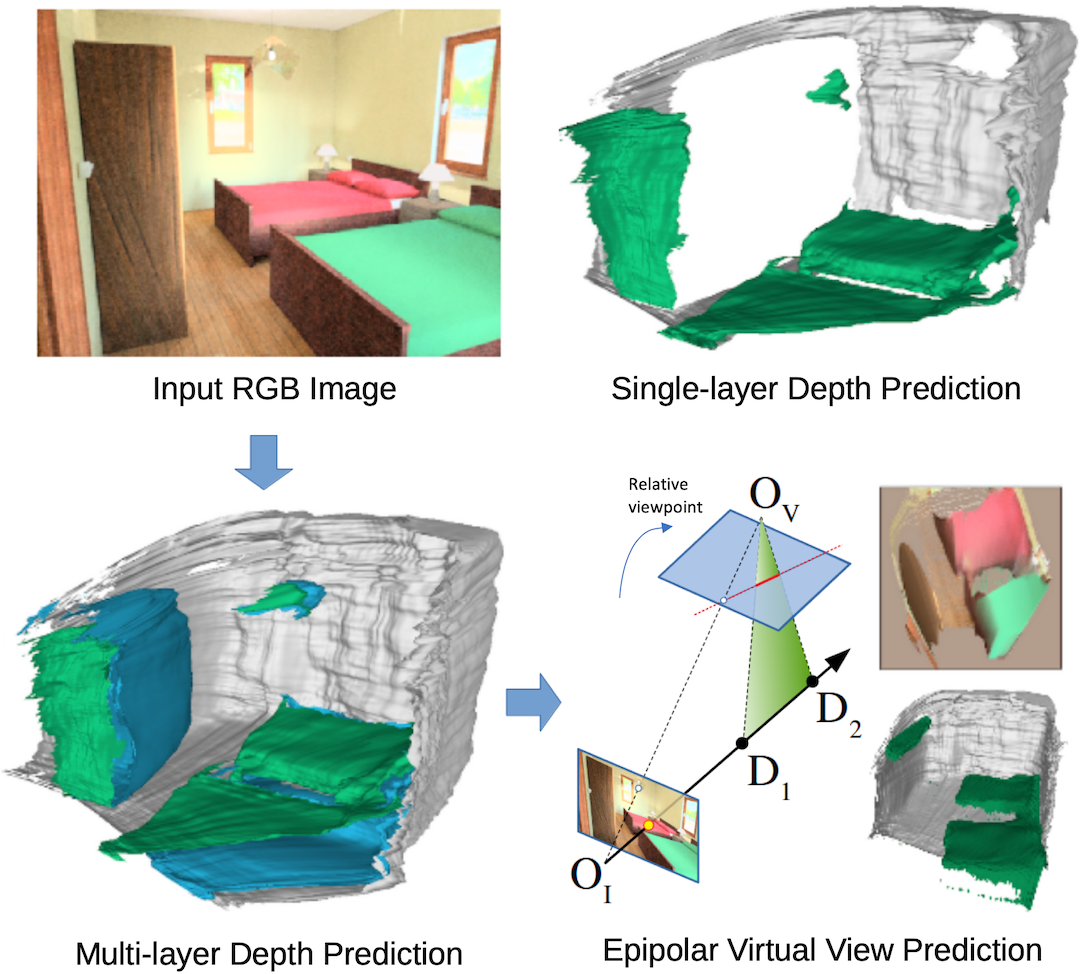}
\end{center}
\vspace{-0.1in}
    \caption{Given a single input view of a scene (top left), we would
    like to predict a complete geometric model. Depth maps (top right)
    provide an efficient representation of scene geometry but are incomplete,
    leaving large holes (e.g., the wardrobe). We propose multi-layer
    depth predictions (bottom left) that provide complete view-based
    representations of shape, and introduce an epipolar transformer network that allows
    view-based inference and prediction from virtual viewpoints 
    (like overhead views, bottom right).}
\label{fig:splash}
\end{figure}

\begin{figure*}
\centering
\includegraphics[width=2\columnwidth,trim={0cm 0cm 0cm 0cm},clip]{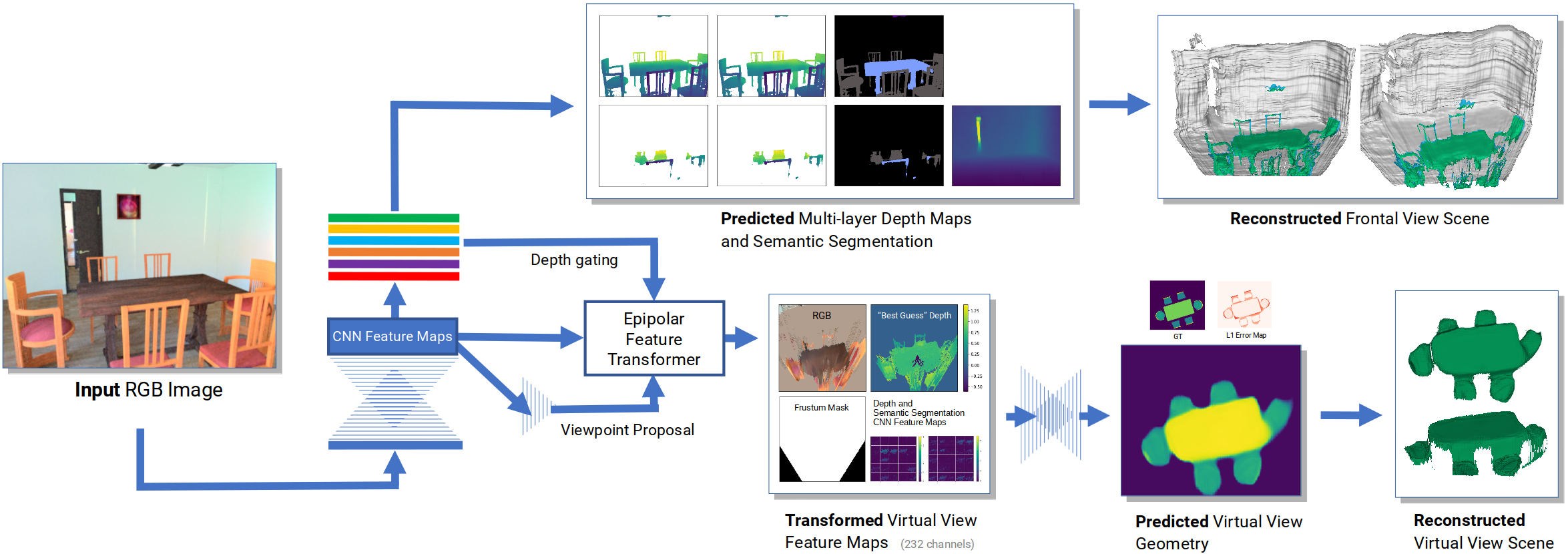}
\caption{
Overview of our system for reconstructing a complete 3D scene from a single RGB
image. We first predict a multi-layer depth map that encodes the
depths of front and back object surfaces as seen from the input
camera. Given the extracted feature map and predicted multi-layer depths, 
the epipolar feature transformer network transfers features from the input view
to a virtual overhead view, where the heights of observed objects are predicted.
Semantic segmentation masks are inferred and inform our geometry estimates, 
but explicit detection of object instances is not required, increasing robustness.}
\label{fig:etn_overview}
\vspace{-0.075in}
\end{figure*}

Early successes at using CNNs for shape prediction leveraged direct
correspondences between the input and output domain, regressing depth and
surface normals at every input pixel \cite{eigen2014depth}. However, these
so-called 2.5D representations are incomplete: they don't make predictions
about the back side of objects or other occluded surfaces.  Several recent
methods instead manipulate voxel-based representations \cite{song2016ssc} and
use convolutions to perform translation-covariant computations in 3D.  This
provides a more complete representation than 2.5D models, but suffers from
substantial storage and computation expense that scales cubically with
resolution of the volume being modeled (without specialized representations
like octrees~\cite{riegler2017octnet}). Other approaches represent shape as
an unstructured point cloud \cite{qi2017pointnet,su18splatnet}, but require
development of suitable convolutional operators
\cite{Gadelha2018multiresolution,wang2018pixel2mesh} and fail to capture
surface topology.

In this paper, we tackle the problem of automatically reconstructing a {\em complete}
3D model of a scene from a single RGB image.  As depicted in
Figures~\ref{fig:splash} and~\ref{fig:etn_overview}, our approach uses an alternative shape representation
that extends view-based 2.5D representations to a complete 3D
representation.  We combine {\em multi-layer} depth maps that store the depth to
multiple surface intersections along each camera ray from a given viewpoint,
with {\em multi-view} depth maps that record surface depths from different camera
viewpoints.

While multi-view and multi-layer shape representations have been explored for
single object shape completion, for example by \cite{shin2018pixels}, we argue
that multi-layer depth maps are particularly well suited for representing full
3D scenes.  {\em First}, they compactly capture high-resolution details about
the shapes of surfaces in a large scene.  Voxel-based representations allocate
a huge amount of resources to simply modeling empty space, ultimately limiting
shape fidelity to much lower resolution than is provided by cues like occluding
contours in the input image~\cite{song2016ssc}.  A multi-layer depth map can be
viewed as a run-length encoding of dense representations that stores only
transitions between empty and occupied space.  {\em Second}, view-based depths
maintain explicit correspondence between input image data and scene geometry.
Much of the work on voxel and point cloud representations for single object
shape prediction has focused on predicting a 3D representation in an
object-centered coordinate system. Utilizing such an approach for scenes
requires additional steps of detecting individual objects and estimating their
pose in order to place them back into some global scene coordinate
system~\cite{tulsiani2018factoring}.  In contrast, view-based multi-depth
predictions provide a single, globally coherent scene representation that can
be computed in a ``fully convolutional'' manner from the input image. 

One limitation of predicting a multi-layer depth representation from the input
image viewpoint is that the representation cannot accurately encode the
geometry of surfaces which are nearly tangent to the viewing direction.
In addition, complicated scenes may contain many partially occluded objects that 
require a large number of layers to represent completely. We address 
this challenge by predicting additional (multi-layer) depth maps computed from
virtual viewpoints elsewhere in the scene. To link these predictions from
virtual viewpoints with the input viewpoint, we introduce a novel
\emph{Epipolar Feature Transformer} (EFT) network module.  Given the relative
poses of the input and virtual cameras, we transfer features from a given
location in the input view feature map to the corresponding epipolar line in
the virtual camera feature map. This transfer process is modulated by
predictions of surface depths from the input view in order to effectively
re-project features to the correct locations in the overhead view. 

To summarize our contributions, we propose a view-based, multi-layer depth
representation that enables fully convolutional inference of 3D scene geometry
and shape completion.  We also introduce 
EFT networks that provide geometrically consistent transfer of CNN features
between cameras with different poses, allowing end-to-end training for
multi-view inference.  We experimentally characterize the completeness of these
representations for describing the 3D geometry of indoor scenes, and show that
models trained to predict these representations can provide better recall
and precision of scene geometry than existing approaches based on object
detection.

\section{Related Work}
The task of recovering 3D geometry from 2D images has a rich
history dating to the visionary work of Roberts~\cite{roberts1963machine}.

\boldhead{Monocular Object Shape Prediction} 
Single-view 3D shape reconstruction is challenging because the
output space is under-constrained. Large-scale datasets like
ShapeNet~\cite{chang2015shapenet,wu20153d} facilitate progress in this area,
and recent methods have learned geometric priors for object
categories~\cite{kar2015category,wu2018shapehd}, disentangled primitive shapes
from objects~\cite{girdhar2016learning,zou20173d}, or modeled
surfaces~\cite{hane2017hierarchical,shin2018pixels,zhang2018learning}. Other 
work aims to complete the occluded geometric structure of objects from
a 2.5D image or partial 3D
scan~\cite{rock2015completing,dai2017shape,wu2017marrnet,yang20183d}.  While
the quality of such 3D object reconstructions continues to
grow~\cite{katoneural,wang2018pixel2mesh}, applications are limited by the
assumption that input images depict a single, centered object.

\boldhead{3D Scene Reconstruction}
We seek to predict the geometry of full scenes containing an
unknown number of objects; this task is significantly more challenging than
object reconstruction.  Tulsiani~\etal~\cite{tulsiani2018factoring} factorize
3D scenes into detected objects and room layout by integrating separate methods
for 2D object detection, pose estimation, and object-centered shape prediction.
Given a depth image as input, Song~\etal~\cite{song2016ssc} propose a
volumetric reconstruction algorithm that predicts semantically labeled 3D
voxels.  Another general approach is to retrieve exemplar CAD models from a
large database and reconstruct parts of
scenes~\cite{izadinia2017im2cad,zou2017complete,gupta2015aligning}, but the
complexity of CAD models may not match real-world environments.  While our
goals are similar to Tulsiani~\etal, our multi-layered depth estimates provide
a denser representation of complex scenes.

\boldhead{Representations for 3D Shape Prediction}
Most recent methods use voxel representations to reconstruct 3D geometry~\cite{choy20163dr2n2,song2016ssc,Edward173D,drcTulsiani17,smith2018multi},
in part because they easily integrate with 3D CNNs~\cite{wu20153d} for
high-level recognition tasks~\cite{maturana2015voxnet}.
Other methods~\cite{fan2017point,lin2018learning} use dense point clouds representations.  Classic 2.5D depth
maps~\cite{eigen2014depth,chen2016single} recover the geometry of visible scene
features, but do not capture occluded regions.  Shin \etal~\cite{shin2018pixels}
empirically compared these representations for object reconstruction.  We
extend these ideas to whole scenes via a multi-view,
multi-layer depth representation that encodes the shape of multiple
objects.

\boldhead{Learning Layered Representations}
Layered representations~\cite{wang94} have proven useful for many computer
vision tasks including segmentation~\cite{ghosh2012nonparametric} and optical
flow prediction~\cite{sun2012layered}.  For 3D reconstruction, decomposing
scenes into layers enables algorithms to reason about object occlusions and
depth orderings~\cite{isola2013scene,smith2004layered,wang2019geometric}. Layered 2.5D
representations such as the two-layer decompositions
of~\cite{tulsiani2018layer,dhamo2018peeking} infer the depth of occluded
surfaces facing the camera. Our multi-layer depth representation extends
this idea by including the depth of back surfaces (equiv. object thickness).
We also infer depths from virtual viewpoints far from the input view
for more complete coverage of 3D scene geometry.  Our use of layers generalizes~\cite{richter2018matryoshka}, who used multiple intersection depths to model
non-convexities for constrained scenes containing a single, centered object.
Concurrently to our work, \cite{nicastro2019x} predicts object-level thicknesses for volumetric RGB-D fusion
and \cite{gabeur2019moulding} estimates 3D human shape.

\boldhead{Multi-view Shape Synthesis} 
Many classic 3D reconstruction methods utilize multi-view inputs to synthesize
3D shapes~\cite{izadi2011kinectfusion,snavely2008modeling,dai2017scannet}.
Given monocular inputs, several recent methods explore ways of synthesizing
object appearance or image features from novel
viewpoints~\cite{zhou2018stereo,yan2016perspective,ji2017deep,choy20163dr2n2,tvsn_cvpr2017,su20143d}.
Other work uses unsupervised learning from stereo or video sequences to reason
about depths~\cite{zhou2017unsupervised,jiang2017self}.  Instead of simply
transferring the pixel colors associated with surface points to novel views,
we transfer whole CNN feature maps over corresponding object volumes, 
and thereby produce more accurate and complete 3D reconstructions.

\section{Reconstruction with Multi-Layer Depth}

\begin{figure}[t]
\begin{center}
\vspace{-0.03in}
  \includegraphics[width=0.9\columnwidth]{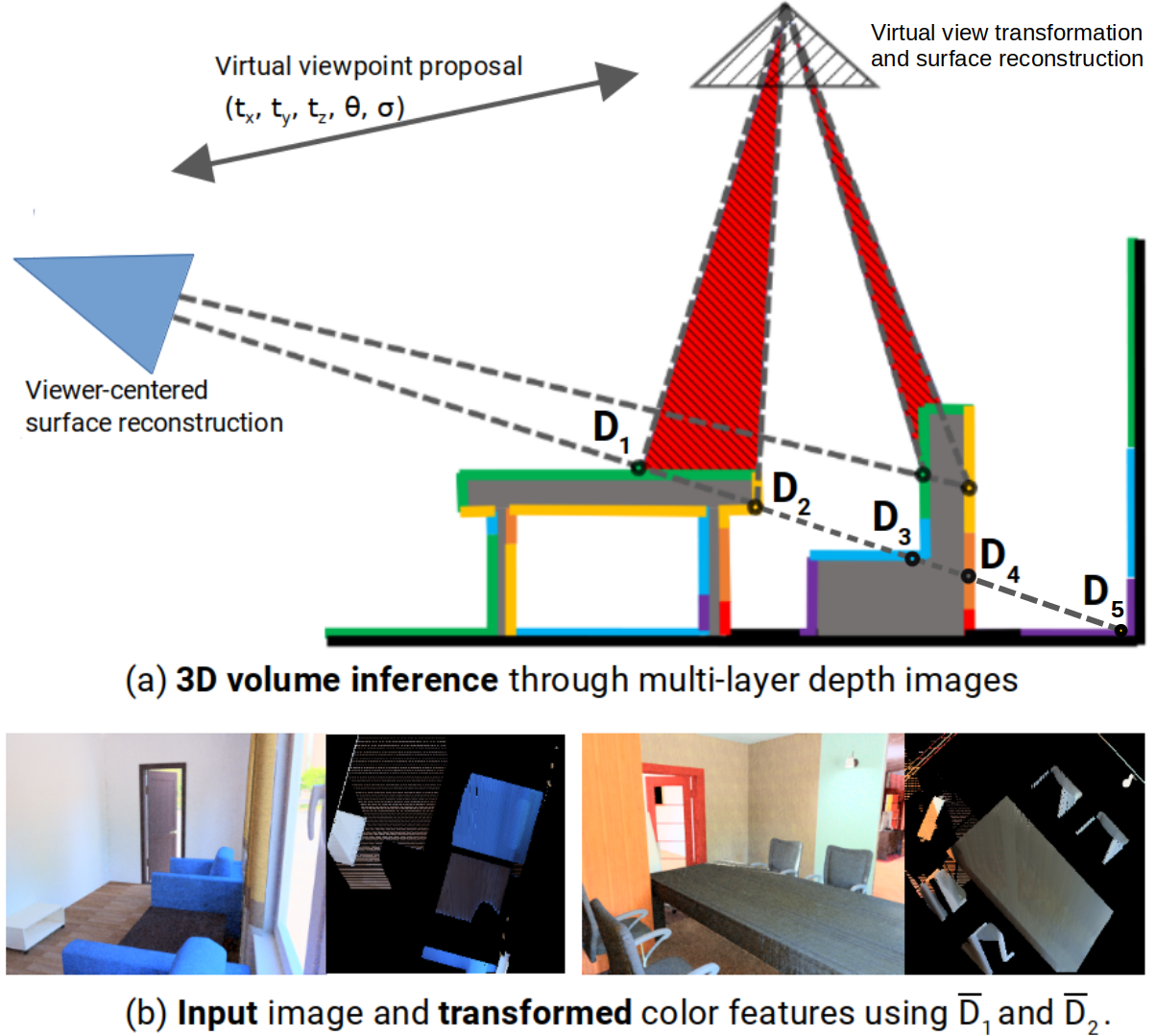}
\end{center}
\vspace{-0.19in}
    \caption{
    Epipolar transfer of features from the input image to a virtual overhead view. 
    Given multi-layer depth predictions of surface entrances and exits,
    each pixel in the input view is mapped to zero, one, or two segments of the
    corresponding epipolar line in the virtual view.}
\label{fig:volumeinference}
\end{figure}

Traditional depth maps record the depth at which a ray through a given pixel
first intersects a surface in the scene. Such 2.5D representations of
scene geometry accurately describe visible surfaces,
but cannot encode the shape of partially occluded objects,
and may fail to capture the complete 3D shape of unoccluded objects
(due to self-occlusion). We instead represent 3D scene geometry 
by recording multiple surface intersections for each camera ray.  As illustrated in
Figure~\ref{fig:volumeinference}(a), some rays may intersect many 
object surfaces and require several layers to capture all details.
But as the number of layers grows, multi-layer depths completely represent 3D scenes with multiple non-convex objects.

\setlength\tabcolsep{3pt} 
\newcolumntype{Y}{>{\arraybackslash\hsize=1.03\hsize}X}
\begin{table}[b]
\vspace{4pt}
\centering
\small
\begin{tabularx}{\columnwidth}{X|X|X|X|X|Y}
$\bar{D}_{1}$  & $\bar{D}_{1,2}$  & $\bar{D}_{1,2,3}$  & $\bar{D}_{1..4}$ & $\bar{D}_{1..5}$ & $\bar{D}_{1..5}\text{ +Ovh.}$    \\ \hline
  0.237  & 0.427 & 0.450 & 0.480 & 0.924 & 0.932 \\
\end{tabularx}
\vspace{-0.125in}
\caption{Scene surface coverage (recall) of ground truth depth layers with
a 5cm threshold.  Our predictions cover 93\% of the scene
geometry inside the viewing frustum.}
\label{fig:ub_table}
\end{table}
\newcolumntype{Y}{>{\centering\arraybackslash\hsize=.5\hsize}X}

We use experiments to empirically determine a fixed number of layers that provides 
good coverage of typical natural scenes, while remaining compact enough for
efficient learning and prediction.  
Another challenge is that surfaces that are nearly tangent to input camera rays
are not well represented by a depth map of fixed resolution.  To address this,
we introduce an additional virtual view where tangent surfaces are sampled more 
densely (see Section \ref{sec:eft}).

\begin{figure}[b]
	\begin{center}
		\includegraphics[width=0.925\columnwidth]{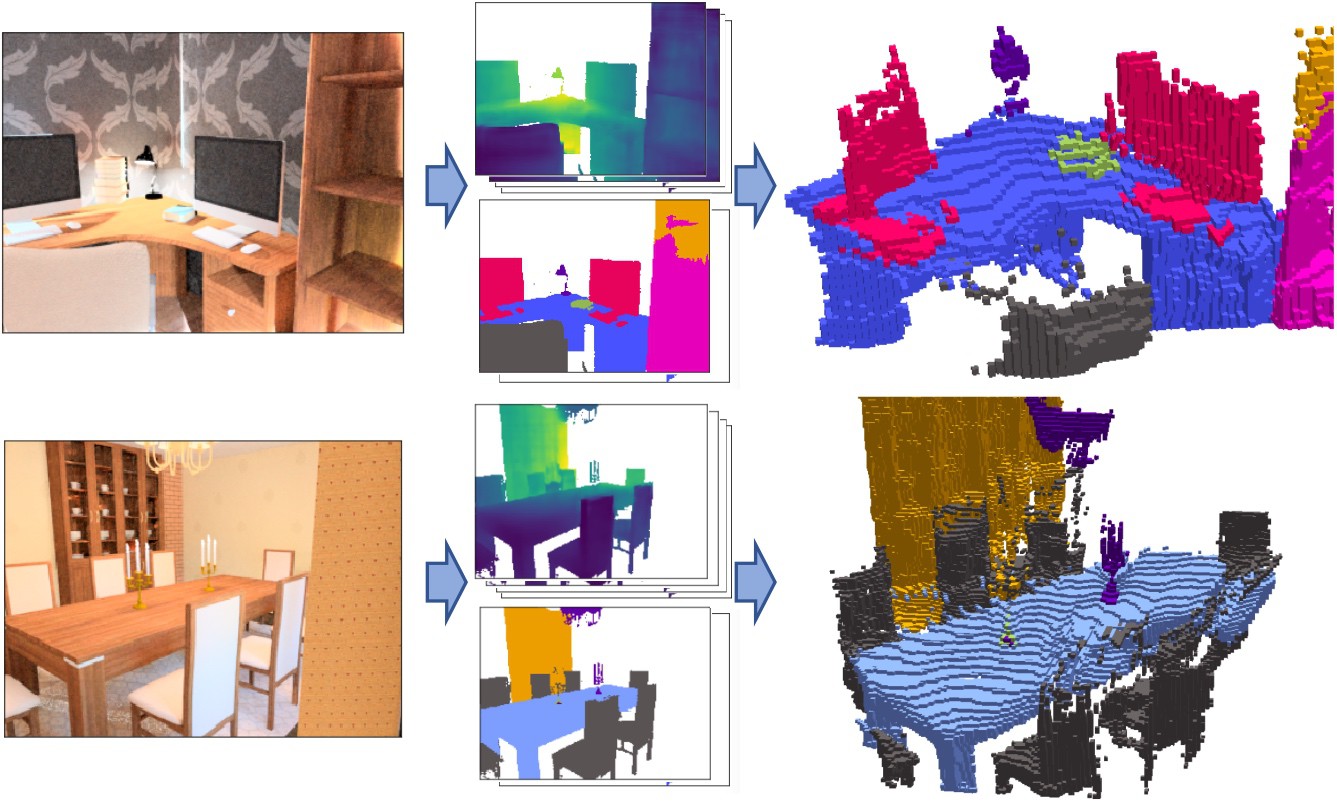}
	\end{center}
	\vspace{-0.22in}
	\caption{
		A volumetric visualization of our predicted multi-layer surfaces and
	    semantic labels on SUNCG.
	    We project the center of each voxel into the input camera, and the voxel is marked occupied if
	    the depth value falls in the first object interval $({D}_1, {D}_2)$ or
	    the occluded object interval $({D}_3, {D}_4)$. 
	}
	\label{fig:voxelization}
\end{figure}

\subsection{Multi-Layer Depth Maps from 3D Geometry}

In our experiments, we focus on a five-layer model designed to represent key
features of 3D scene geometry for typical indoor scenes.  To capture the
overall room layout, we model the room envelope
(floors, walls, ceiling, windows) that defines the extent of the space. We define
the depth $D_5$ of these surfaces to be the \emph{last} layer of the scene.

To model the shapes of observed objects, we trace rays from the input
view and record the first intersection with a visible surface in
depth map $D_1$. This resembles a standard depth map, but excludes the room
envelope. If we continue along the same ray, it will eventually exit the object
at a depth we denote by $D_2$. For non-convex objects the ray may
intersect the same object multiple times, but we only record the \emph{last}
exit in $D_2$.  As many indoor objects have large convex parts, the $D_1$ and
$D_2$ layers are often sufficient to accurately reconstruct a large proportion
of foreground objects in real scenes.  While room envelopes 
typically have a very simple shape, the prediction of occluded structure behind
foreground objects is more challenging.  We define layer $D_3 > D_2$ as the depth 
of the next object intersection, and $D_4$ as the depth of the exit from that second object instance.

We let $(\bar{D}_1, \bar{D}_2, \bar{D}_3, \bar{D}_4,\bar{D}_5)$ denote the
ground truth multi-layer depth maps derived from a complete 3D model.  Since
not all viewing rays intersect the same number of objects (e.g., when the room
envelope is directly visible), we define a binary mask $\bar{M}_\ell$ which
indicates the pixels where layer $\ell$ has support.  
Note that $\bar{M}_1=\bar{M}_2$, and $\bar{M}_3=\bar{M}_4$,
since $D_2$ (first instance exit) has the same support as $D_1$.
Experiments in Section~\ref{sec:experiments} evaluate the relative importance
of different layers in modeling realistic 3D scenes.

\subsection{Predicting Multi-Layer Depth Maps}
\label{sec:predicing}

To learn to predict five-channel multi-layer depths $\mathcal{D}
= (D_1, D_2, D_3, D_4, D_5)$ from images, we train a standard encoder-decoder
network with skip connections, and use the Huber loss $\rho_h(., .)$ to measure
prediction errors:
\begin{equation}
  L_d (\mathcal{D}) = \sum_{\ell=1}^5 
  \left( \frac{\bar{M}_\ell}{||\bar{M}_\ell||_1} \right)
  \cdot \rho_h(D_\ell, \bar{D}_\ell).
\label{eq:loss_func}
\end{equation}
We also predict semantic segmentation masks for the first
and third layers.
The structure of the semantic segmentation network is similar to the multi-layer
depth prediction network, except that the output has 80 channels (40 object
categories in each of two layers), and errors are measured via the cross-entropy loss.
To reconstruct complete 3D geometry from multi-layer depth predictions, we use
predicted masks and depths to generate meshes corresponding to the front and back
surfaces of visible and partially occluded objects, as well as the room envelope.
Without the back surfaces \cite{shade1998layered}, ground truth depth layers $\bar{D}_{1,3,5}$ cover only 82\% of the scene geometry inside the viewing
frustum (vs. 92\% including back surfaces of objects, see Table~\ref{fig:ub_table}).

\section{Epipolar Feature Transformer Networks}
\label{sec:eft}

To allow for richer view-based scene understanding, we would like to relate
features visible in the input view to feature representations in other views.
To achieve this, we transfer features computed in input 
image coordinates to the coordinate system of a ``virtual camera''
placed elsewhere in the scene. This approach more efficiently covers some
parts of 3D scenes than single-view, multi-layer depths.

Figure~\ref{fig:etn_overview} shows a block diagram of our 
\emph{Epipolar Feature Transformer} (EFT) network.  Given features $F$
extracted from the image, we choose a virtual camera location with
transformation mapping $T$ and transfer weights $W$, and use these to warp
$F$ to create a new ``virtual view'' feature map $G$.
Like \emph{spatial transformer networks} (STNs)~\cite{jaderberg2015spatial} 
we perform a parametric, differentiable ``warping'' of a feature map.  However, 
EFTs incorporate a weighted pooling operation informed by multi-view geometry.

\boldhead{Epipolar Feature Mapping}
Image features at spatial location $(s,t)$ in an input view correspond to 
information about the scene which lies somewhere along the ray
\[
 \begin{bmatrix}
  x \\ y \\ z
 \end{bmatrix} 
 = z 
\mathbf{K_I}^{-1}
\begin{bmatrix}
s \\ t \\ 1
\end{bmatrix},
\quad \quad z \geq 0,
\]
where $\mathbf{K}_I \in \mathbb{R}^{3 \times 3}$ encodes the input camera
intrinsic parameters, as well as the spatial resolution and offset of the
feature map. $z$ is the depth along the viewing ray, whose image
in a virtual orthographic camera is given by
\[
 \begin{bmatrix}
  u(s,t,z) \\ v(s,t,z) 
 \end{bmatrix} 
 =
\mathbf{K_V}
\left(
z
\mathbf{R}
\mathbf{K_I}^{-1}
\begin{bmatrix}
s \\ t \\ 1
\end{bmatrix}
+ \mathbf{t}
\right),
\quad \quad z \geq 0.
\]
Here $\mathbf{K}_V \in \mathbb{R}^{2 \times 3}$ encodes the virtual view
resolution and offset, and $\mathbf{R}$ and $\mathbf{t}$ the relative
pose.\footnote{For a perspective model the righthand side is scaled by $z'(s,t,z)$, the
depth from the virtual camera of the point at location $z$ along the ray.}  Let
$T(s,t,z) = (u(s,t,z),v(s,t,z))$ denote the forward mapping from points along
the ray into the virtual camera, and $\Omega(u,v) = \{ (s,t,z) : T(s,t,z) =
(u,v) \}$ be the pre-image of $(u,v)$. 

Given a feature map computed from the input view $F(s,t,f)$, where $f$ indexes
the feature dimension, we synthesize a new feature map $G$
corresponding to the virtual view.  We consider general mappings of the form
\[
G(u,v,f) = \frac{\sum_{(s,t,z) \in \Omega(u,v)} F(s,t,f) W(s,t,z)}{\sum_{(s,t,z) \in \Omega(u,v)} W(s,t,z)},
\]
where $W(s,t,z) \geq 0$ is a gating function that may depend on features of the input
image.\footnote{For notational simplicity, we have written $G$ as a sum over
a discrete set of samples $\Omega$.  To make $G$ differentiable with respect to
the virtual camera parameters, we perform bilinear interpolation.}
When $\Omega(u,v)$ is empty, we set $G(u,v,f)=0$ for points $(u,v)$ outside the viewing
frustum of the input camera, and otherwise interpolate feature values from those
of neighboring virtual-view pixels.

\begin{figure}
\centering
\begin{tabular}{ccc}
\subfigimg[width=0.30\columnwidth]{Input}{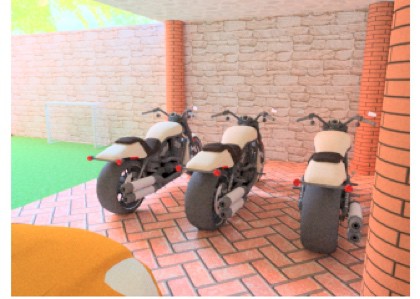}&
\subfigimg[width=0.30\columnwidth]{3D Reconstruction}{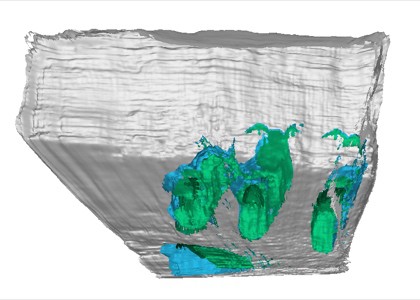}&
\subfigimg[width=0.30\columnwidth]{3D Reconstruction}{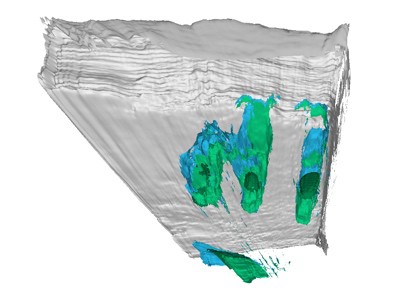}\\
\subfigimg[width=0.30\columnwidth]{Input}{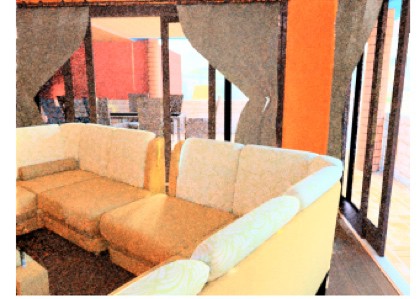}&
\subfigimg[width=0.30\columnwidth]{3D Reconstruction}{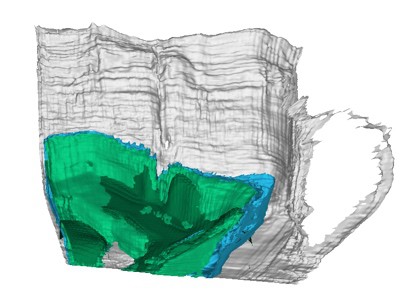}&
\subfigimg[width=0.30\columnwidth]{3D Reconstruction}{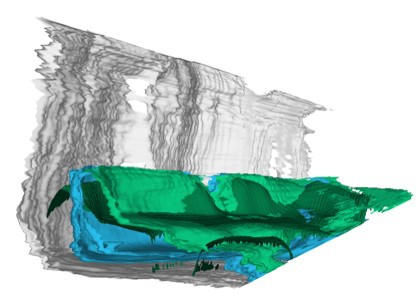}\\
\subfigimg[width=0.30\columnwidth]{Input}{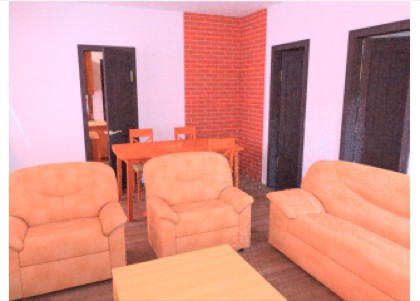}&
\subfigimg[width=0.30\columnwidth]{3D Reconstruction}{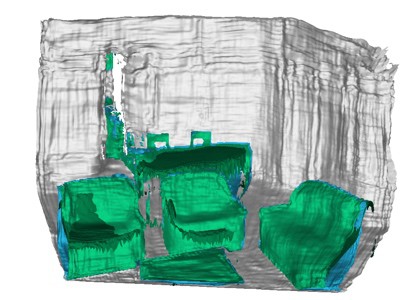}&
\subfigimg[width=0.30\columnwidth]{3D Reconstruction}{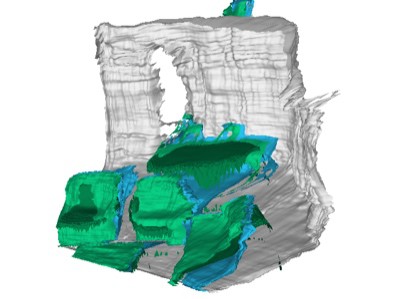}\\
\subfigimg[width=0.30\columnwidth]{Input}{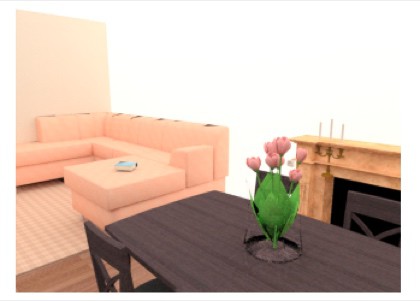}&
\subfigimg[width=0.30\columnwidth]{3D Reconstruction}{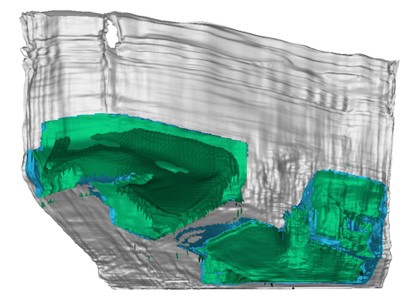}&
\subfigimg[width=0.30\columnwidth]{3D Reconstruction}{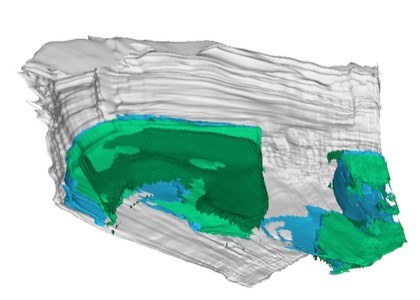}\\
\end{tabular}
\vspace{-0.11in}
\caption{Single image scene reconstruction via multi-layer depth maps.
  Estimates of the front (green) and back (cyan) surfaces of objects, as seen from the input view, are complemented by heights estimated by a virtual overhead camera (dark green) via our epipolar feature transform.  Room envelope estimates are rendered in gray.}
\label{fig:exp_recon}
\end{figure}

\boldhead{Choice of the Gating Function \boldmath{$W$}}
By design, the transformed features are differentiable with respect to $F$ and $W$.
Thus in general we could assign a loss to predictions from the virtual camera,
and learn an arbitrary gating function $W$ from training data. 
However, we instead propose to leverage additional geometric structure based on 
predictions about the scene geometry produced by the frontal view.

Suppose we have a scene depth estimate $D_1(s,t)$ at every location 
in the input view.  To simplify occlusion reasoning we assume
that relative to the input camera view, the virtual camera is rotated around
the $x$-axis by $\theta < 90^{\circ}$ and translated in $y$ and $z$ to sit above the
scene so that points which project to larger $t$ in the input view have larger
depth in the virtual view. Setting the gating function as
\begin{small}
\[
  W^1_{\text{surf}}(s,t,z) = \delta[D_1(s,t) = z]  \prod_{\hat t=0}^{t-1} \delta[D_1( s,\hat t) + (t - \hat t) \cos \theta \not= z]
\]
\end{small}
\!\!yields an epipolar feature transform that {\em re-projects} each feature at
input location $(s,t)$ into the overhead view via the depth 
estimate $D_1$, but only in cases where it is not occluded by a patch of surface higher up in
the scene. In our experiments we compute $W^\ell_{\text{surf}}$ for each $D_\ell$,
$\ell \in \{1,2,3,4\}$,
and use $W_{\text{surf}} = \max_\ell W^\ell_{\text{surf}}$ to transfer input view features
to both visible and occluded surfaces in the overhead feature map. 
We implement this transformation using a z-buffering approach by traversing 
the input feature map from bottom to top, and overwriting cells in the overhead feature map. 

Figure~\ref{fig:volumeinference}(b) illustrates this feature mapping applied
to color features using the ground-truth depth map for a scene. In some sense,
this surface-based reprojection is quite conservative because it leaves holes in
the interior of objects (e.g., the interior of the orange wood cabinet). If the
frontal view network features at a given spatial location encode the presence,
shape, and pose of some object, then those features really describe a whole
volume of the scene behind the object surface.  It may thus be preferable to
instead transfer the input view features to the entire expected volume 
in the overhead representation.

To achieve this, we use the multi-layer depth representation predicted by the 
frontal view to define a range of scene depths to which the
input view feature should be mapped.  If $D_1(s,t)$ is the depth of the front
surface and $D_2(s,t)$ is the depth at which the ray exits the back surface of
an object instance, we define a volume-based gating function:
\[
  W_{\text{vol}}(s,t,z) = \delta[z \in (D_1(s,t),D_2(s,t))].
\]
As illustrated in Figure~\ref{fig:volumeinference}(a), volume-based gating
copies features from the input view to entire segments
of the epipolar line in the virtual view. In our experiments we use this
gating to generate features for $(D_1,D_2)$ and concatenate them with 
a feature map generated using $(D_3,D_4)$.

\begin{figure}
\centering
\begin{tabular}{ccc}
\subfigimg[width=0.30\columnwidth]{Input}{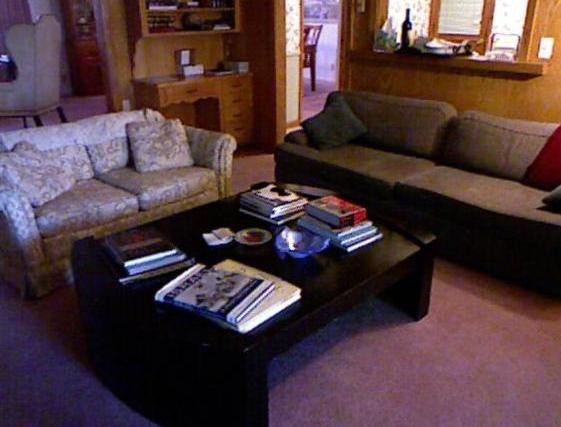}&
\subfigimg[width=0.30\columnwidth]{[Tulsiani, 2018]}{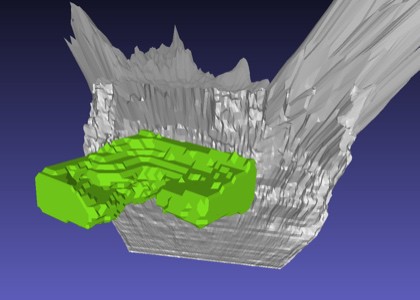}&
\subfigimg[width=0.30\columnwidth]{Ours}{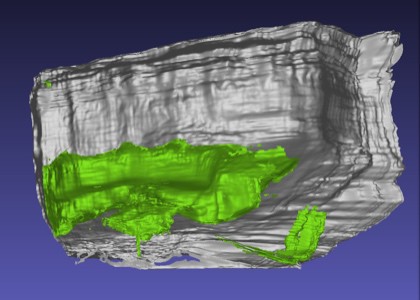}\\
\subfigimg[width=0.30\columnwidth]{Input}{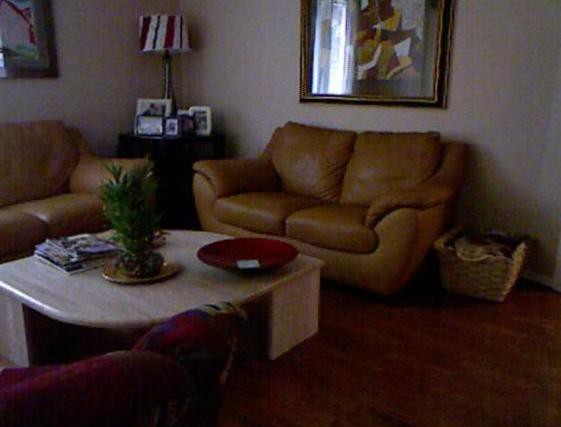}&
\subfigimg[width=0.30\columnwidth]{[Tulsiani, 2018]}{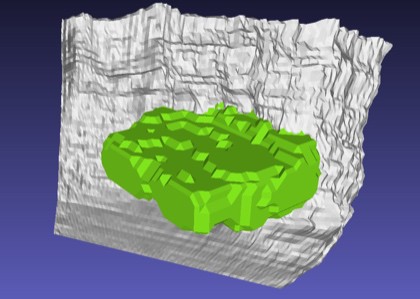}&
\subfigimg[width=0.30\columnwidth]{Ours}{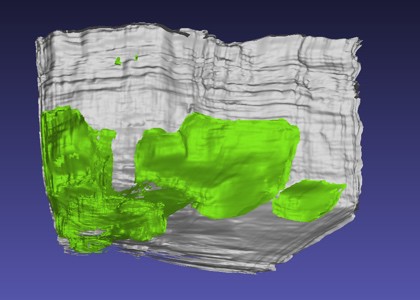}\\
\subfigimg[width=0.30\columnwidth]{Input}{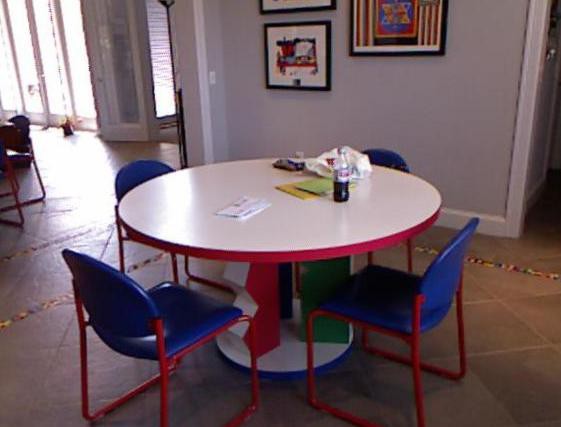}&
\subfigimg[width=0.30\columnwidth]{[Tulsiani, 2018]}{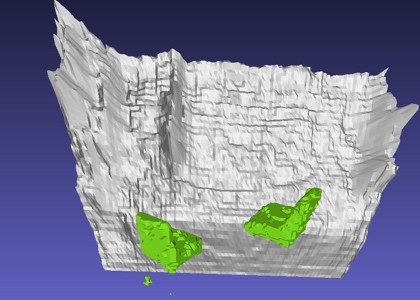}&
\subfigimg[width=0.30\columnwidth]{Ours}{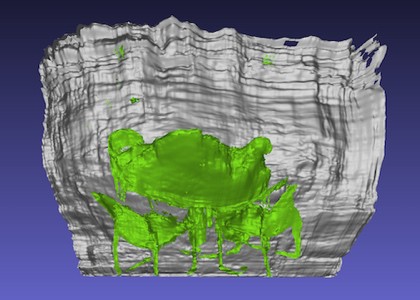}\\
\subfigimg[width=0.30\columnwidth]{Input}{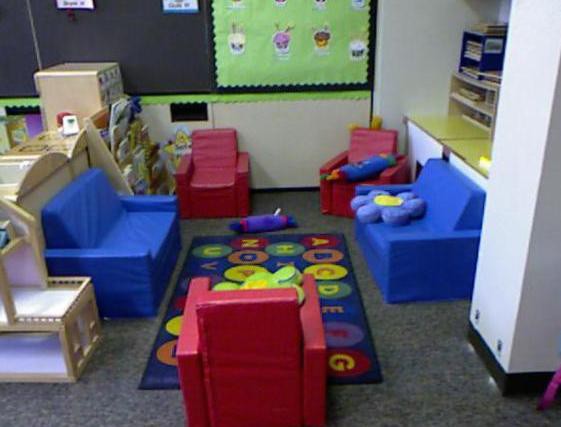}&
\subfigimg[width=0.30\columnwidth]{[Tulsiani, 2018]}{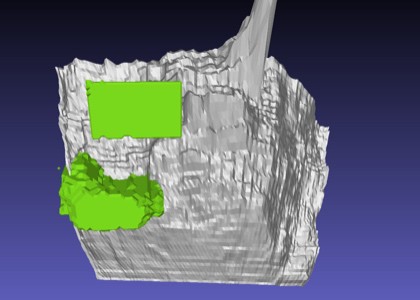}&
\subfigimg[width=0.30\columnwidth]{Ours}{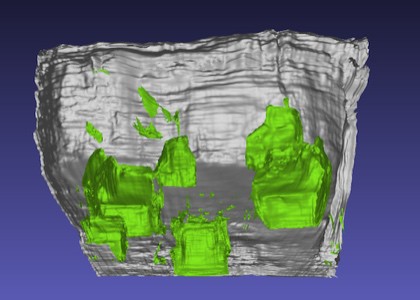}\\
\end{tabular}
\vspace{-0.1in}
\caption{Evaluation of 3D reconstruction on the NYUv2~\cite{SilbermanECCV12}
  dataset. Tulsiani~\etal~\cite{tulsiani2018factoring}
are sensitive to the performance of 2D object detectors, and their voxelized output
  is a coarse approximation of the true 3D geometry.}
\label{fig:nyu}
\end{figure}

\boldhead{Overhead Viewpoint Generation}
For cluttered indoor scenes, there may be many overlapping objects in the input
view. Overhead views of such scenes typically have
much less occlusion and should be simpler to reason about geometrically.  We thus
select a virtual camera that is roughly overhead and covers the
scene content visible from the reference view.  We assume the input view is
always taken with the gravity direction in the $y,z$ plane.  We parameterize the
overhead camera relative to the reference view by a translation $(t_x,t_y,t_z)$
which centers it over the scene at fixed height above the floor, a rotation
$\theta$ which aligns the overhead camera to the gravity direction, and a scale
$\sigma$ that captures the radius of the orthographic camera frustum.

\newcolumntype{Y}{>{\centering\arraybackslash\hsize=.5\hsize}X}
\begin{table}[b]
\vspace{4pt}
	\includegraphics[width=0.92\columnwidth]{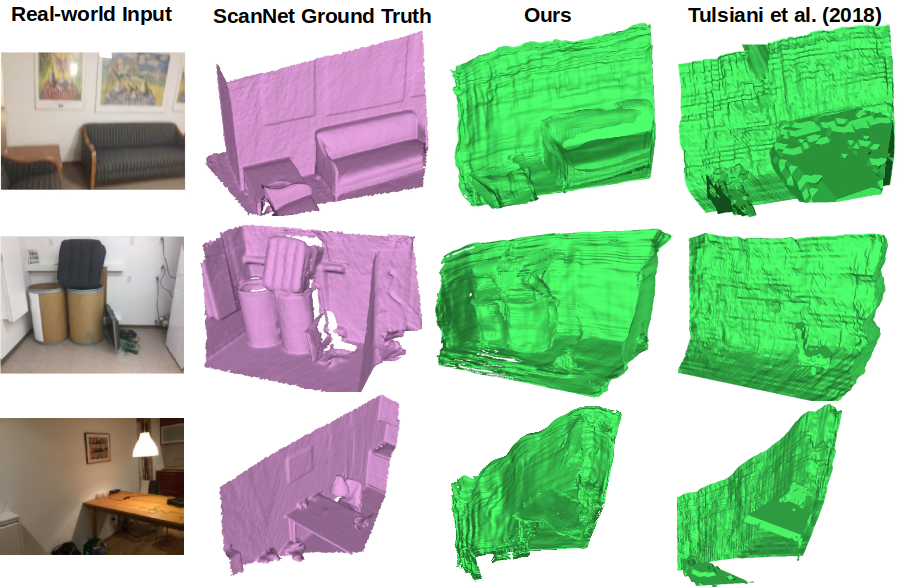}
\centering
\vspace{-1.5pt}
\begin{tabularx}{\columnwidth}{@{}XYY@{}}
 &     Precision & Recall\\ 
\hline
  $D_{1,2,3,4,5}$ \& Overhead &    \textbf{0.221}    & \textbf{0.358}     \\
  Tulsiani~\etal~\cite{tulsiani2018factoring}  &                0.132    & 0.191     \\ 
\end{tabularx}
\vspace{-4.5pt}
\caption{We quantitatively evaluate the synthetic-to-real transfer of 3D
  geometry prediction on the ScanNet~\cite{dai2017scannet} dataset (threshold of 10cm).
We measure recovery of true object surfaces and room layouts within the viewing frustum.
}
\label{fig:scannet_table}
\end{table}

\section{Experiments}
\label{sec:experiments}

Because we model complete descriptions of the ground-truth 3D geometry
corresponding to RGB images, which is not readily available for natural images, 
we learn to predict multi-layer and multi-view depths 
from physical renderings of indoor scenes~\cite{zhang2016physically} 
provided by the SUNCG dataset~\cite{song2016ssc}.

\subsection{Generation of Training Data}
\label{sec:trainingData}
The SUNCG dataset~\cite{song2016ssc} contains complete 3D meshes for 41,490 
houses that we render to generate our training data. 
For each rendered training image, we extract the subset of the house model 
that is relevant to the image content, without making assumptions about the room size.
We transform the house mesh to the camera's coordinate system
and truncate polygons that are outside the left, top, right, bottom, and near
planes of the perspective viewing frustum. Objects that are projected behind
the depth image of the room envelope are also removed. 
The final ground truth mesh that we evaluate against contains all polygons 
from the remaining objects, as well as the true room envelope.

For each rendered training image, we generate target
multi-depth maps and segmentation masks by performing multi-hit ray tracing on
the ground-truth geometry. We similarly compute ground-truth height maps 
for a virtual orthographic camera centered over each scene.
To select an overhead camera viewpoint for training that covers the relevant
scene content, we consider three heuristics: (i)
Convert the true depth map to a point cloud, center the overhead camera over the 
mean of these points, and set the camera radius to 1.5 times their standard deviation;
(ii) Center the overhead camera so that its principal axis
lies in the same plane as the input camera, and offset in front of the
input view by the mean of the room envelope depths; (iii) Select a square
bounding box in the overhead view that encloses all points belonging to objects
visible from the input view. None of these heuristics worked perfectly for all 
training examples, so we compute our final overhead camera view via a weighted average of these three candidates.

\subsection{Model Architecture and Training} 
As illustrated in Figure~\ref{fig:etn_overview}, given an RGB image, we first 
predict a multi-layer depth map as well as a 2D semantic segmentation map. 
Features used to predict multi-layer depths are then mapped through our EFT network
to synthesize features for a virtual camera view, and predict an orthographic height map.
We then use the multi-layer depth map, semantic segmentation map, 
and overhead height map to predict a dense 3D reconstruction of the imaged scene.

\begin{figure}[t]
\begin{center}
\vspace{-0.03in}
  \includegraphics[width=1\columnwidth]{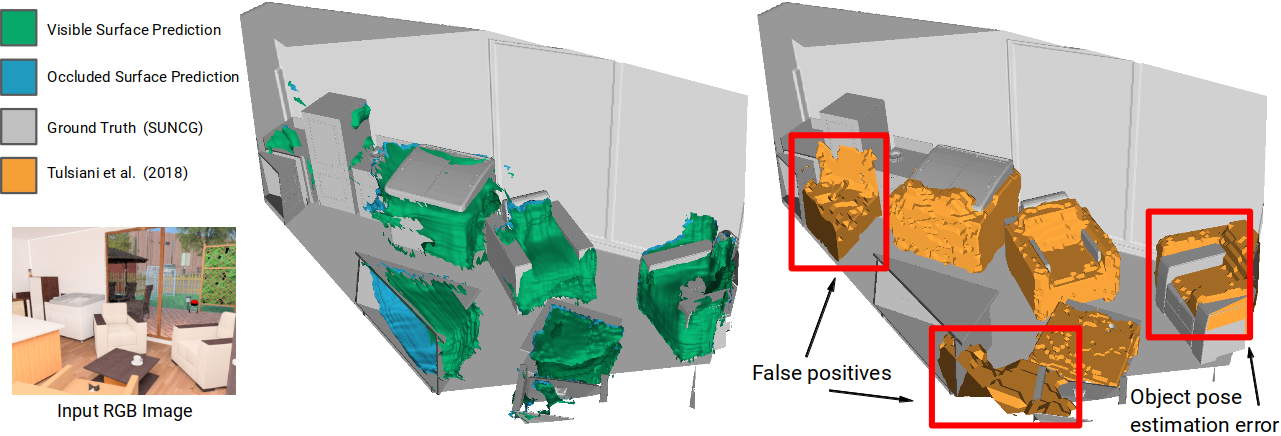}
\end{center}
\vspace{-0.21in}
    \caption{
    Qualitative comparison of our viewer-centered, end-to-end scene surface
    prediction (left) and the object-based detection and voxel shape prediction
    of ~\cite{tulsiani2018factoring} (right).  Object-based reconstruction is
    sensitive to detection and pose estimation errors, while our method is more robust.
    }
\label{fig:soa}
\end{figure}

We predict multi-layer depth maps and semantic segmentations 
via a standard convolutional encoder-decoder with skip connections. The network
uses dilated convolution and has separate output branches for predicting each
depth layer using the Huber loss specified in Section~\ref{sec:predicing}. For
segmentation, we train a single branch network using a softmax loss
to predict 40 semantic categories derived from the SUNCG mesh labels
(see supplement for details).

Our overhead height map prediction network takes as input the transformed
features of our input view multi-layer depth map. The overhead model integrates 
232 channels (see Figure~\ref{fig:etn_overview}) including epipolar
transformations of a $48$-channel feature map from the depth prediction network, 
a $64$-channel feature map from the semantic segmentation network, and the
RGB input image. These feature maps are extracted from the frontal networks
just prior to the predictive branches.  Other inputs include a ``best
guess'' overhead height map derived from frontal depth predictions, and a 
mask indicating the support of the input camera frustum.
The frustum mask can be computed by applying the epipolar transform
with $F=1$, $W=1$. The best-guess overhead depth map can be computed 
by applying an unnormalized gating function $W(s,t,z) = z \cdot
\delta[D_1(s,t)=z]$ to the $y$-coordinate feature $F(s,t)=t$.

We also train a model to predict the virtual camera parameters which takes
as input feature maps from our multi-depth prediction network, 
and attempts to predict the target overhead viewpoint (orthographic translation $(t_x,t_y)$ and frustum radius $\sigma$)
chosen as in Section~\ref{sec:trainingData}.   
While the EFT network is differentiable and our final model can in
principle be trained end-to-end, in our
experiments we simply train the frontal model to convergence, freeze it, and
then train the overhead model on transformed features without backpropagating
overhead loss back into the frontal-view model parameters. We use the Adam
optimizer to train all of our models with batch size 24 and learning rate
0.0005 for 40 epochs.  The Physically-based
Rendering~\cite{zhang2016physically} dataset uses a fixed downward tilt camera
angle of 11 degrees, so we do not need to predict the gravity angle. 
At test time, the height of the virtual camera is the same as the input frontal 
camera and assumed to be known. We show qualitative 3D reconstruction results on the SUNCG test set in Figure~\ref{fig:exp_recon}.

\begin{figure}[t]
\centering
\includegraphics[trim={0 0 0 0},clip,width=0.49\linewidth]{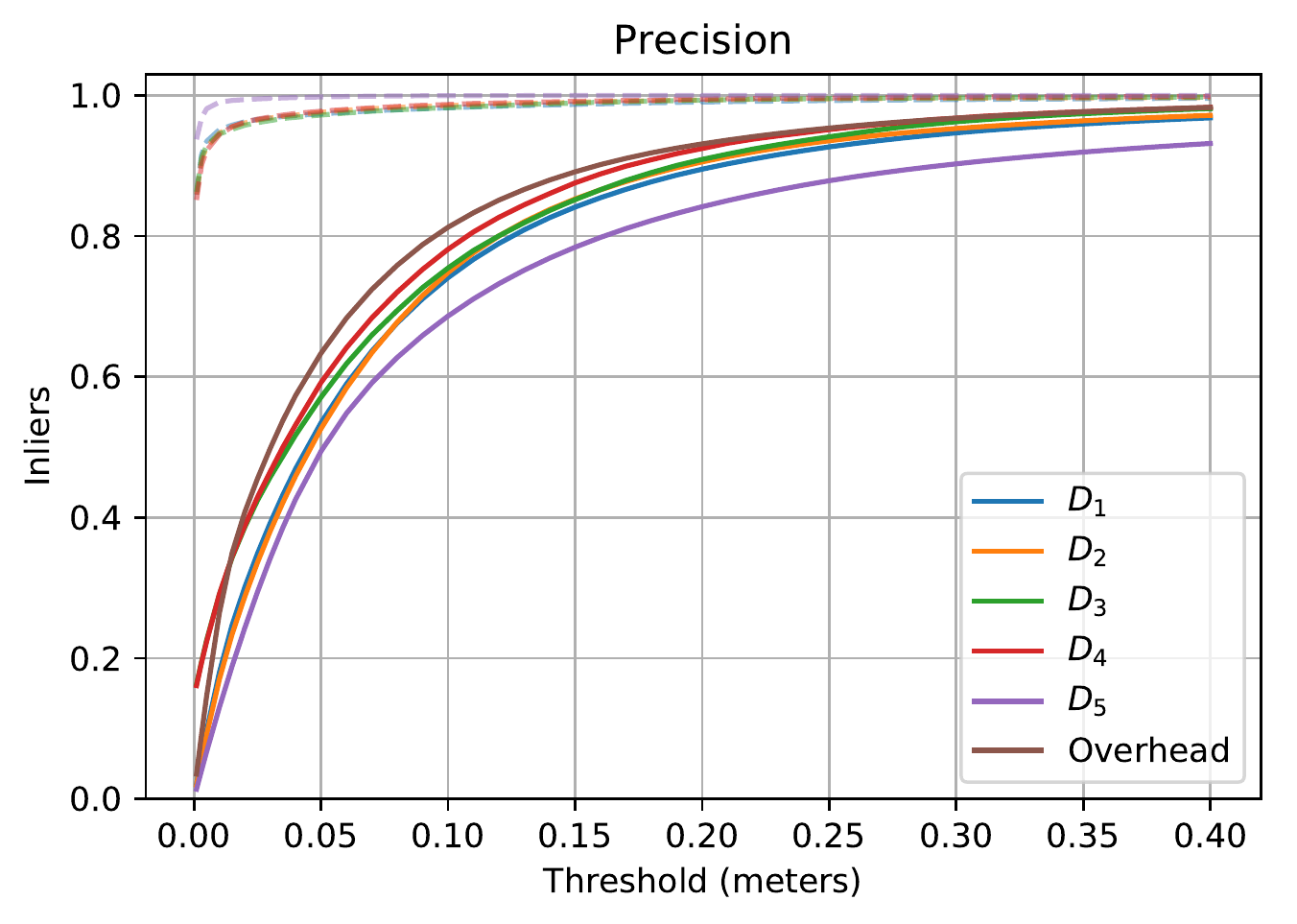}
\includegraphics[trim={0 0 0 0},clip,width=0.49\linewidth]{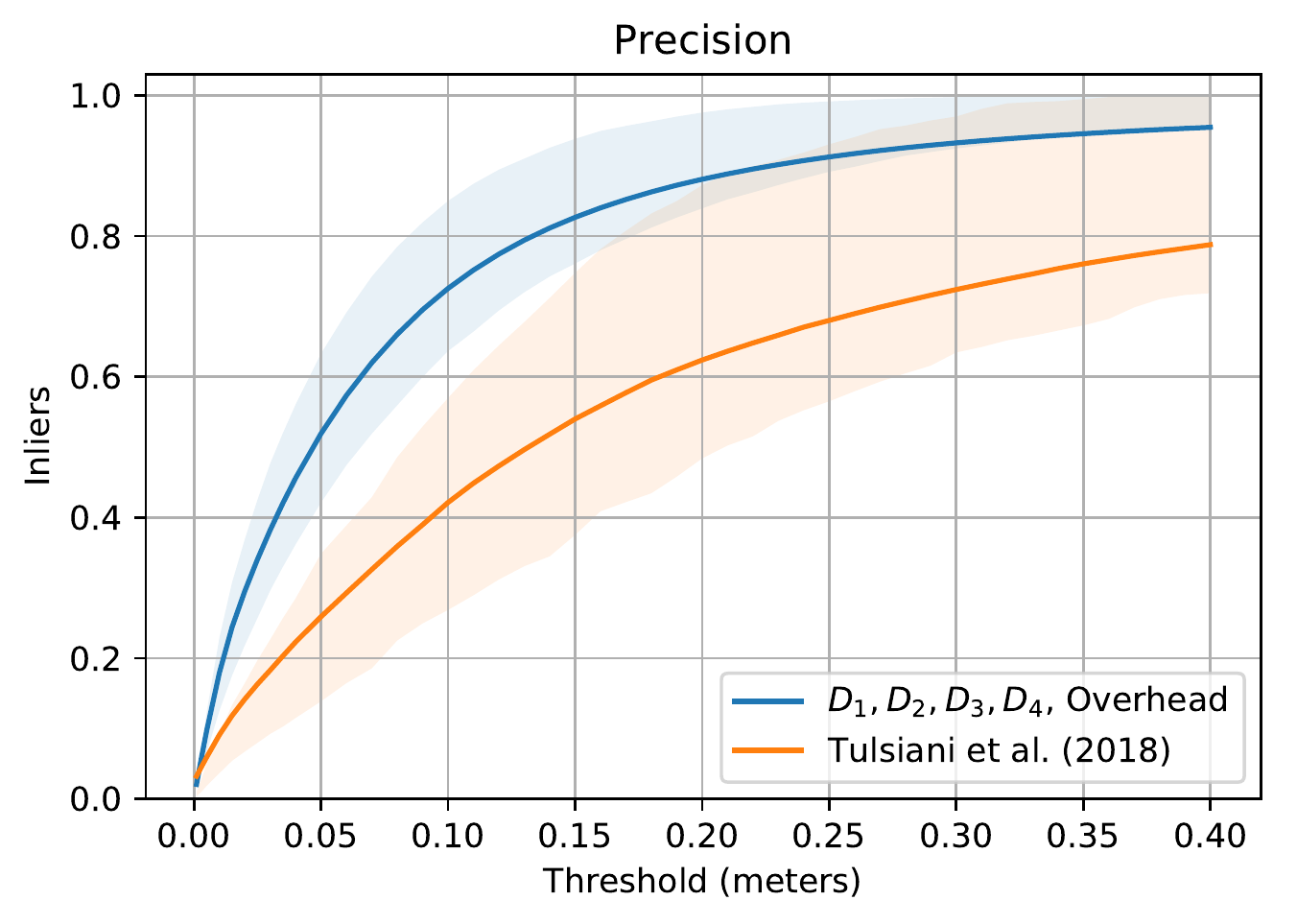}
\includegraphics[trim={0 0 0 0},clip,width=0.49\linewidth]{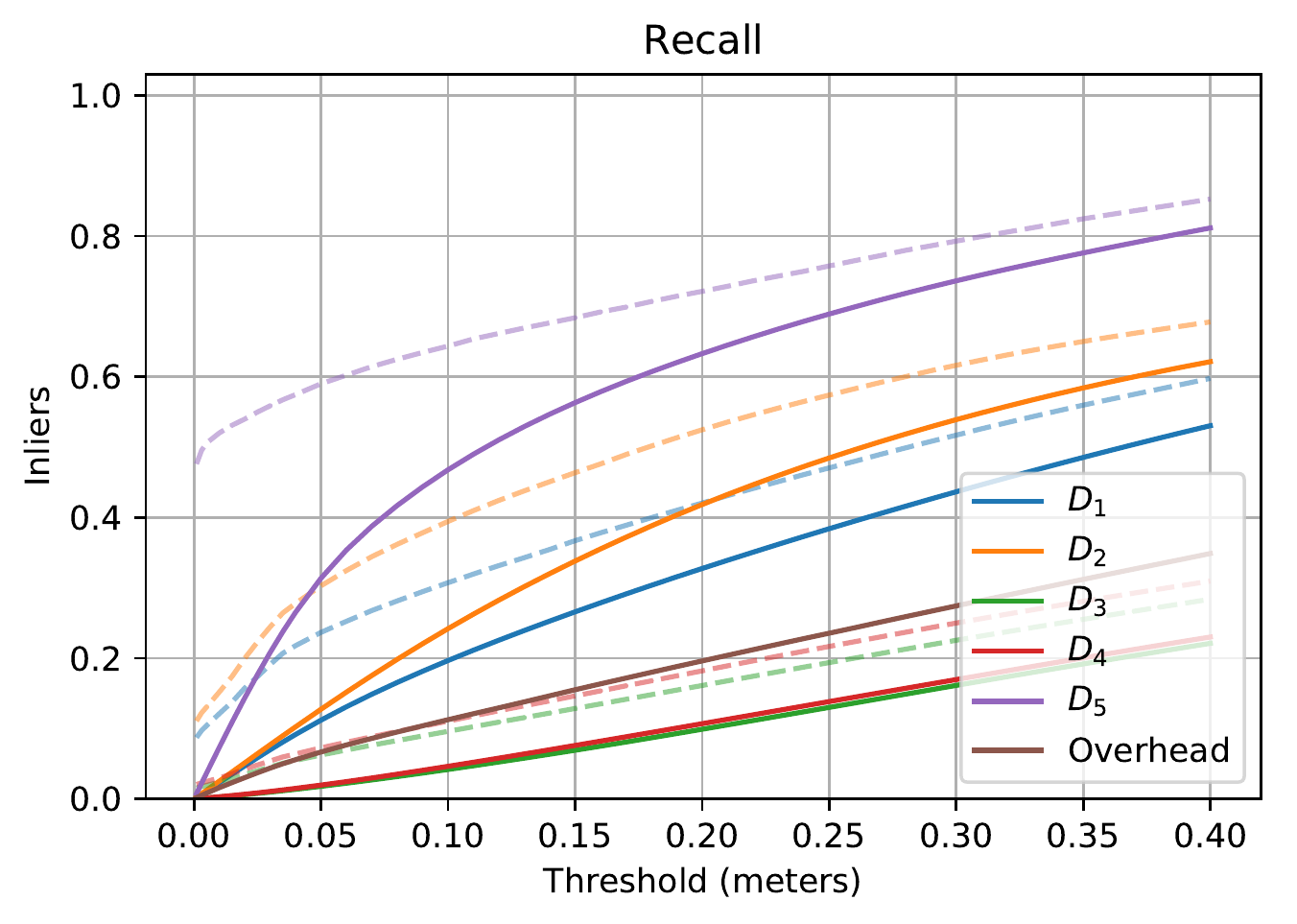}
\includegraphics[trim={0 0 0 0},clip,width=0.49\linewidth]{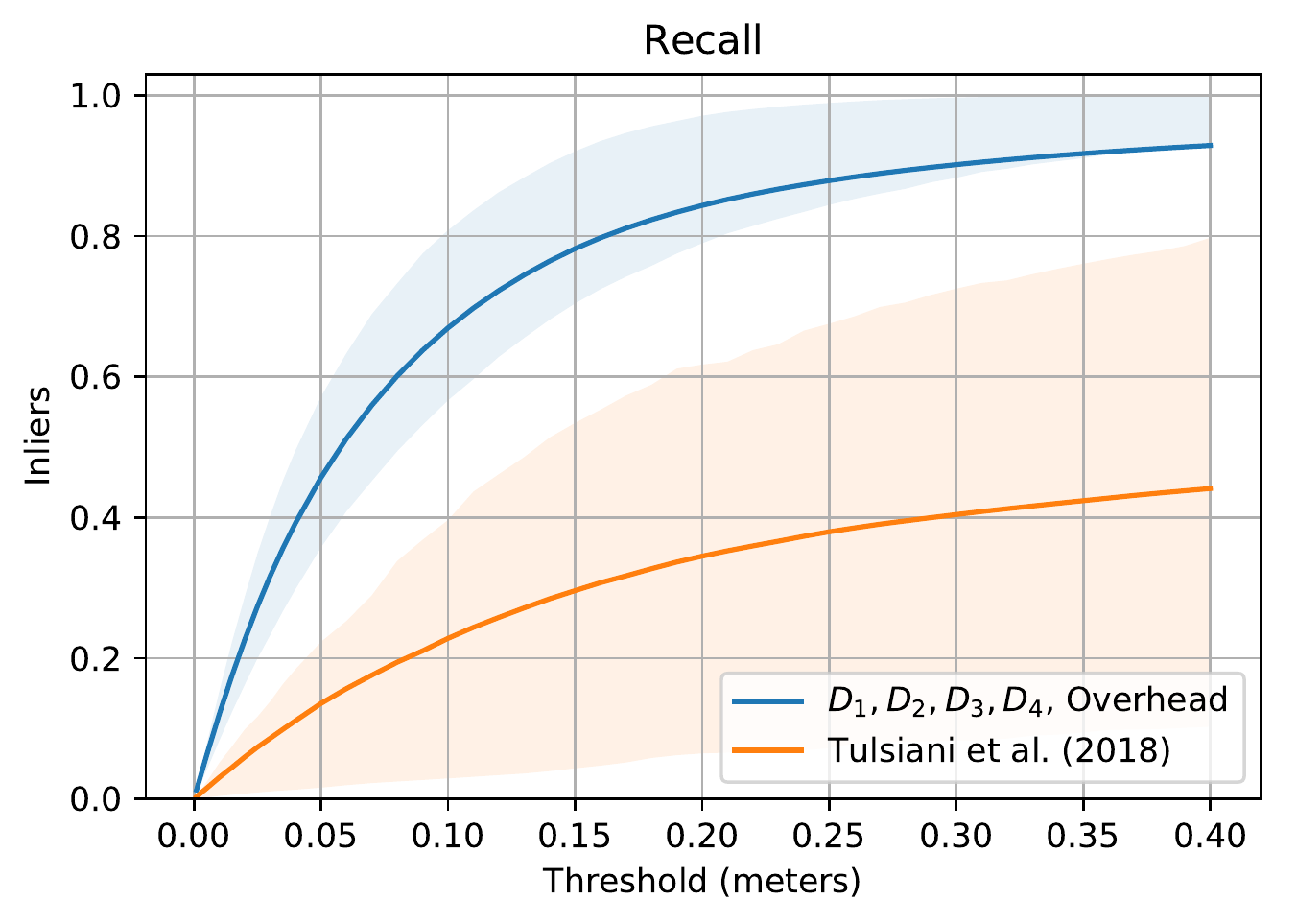}
\vspace{-0.9em}
\caption{
  Precision and recall of scene geometry as a function of match distance threshold.
  \textit{Left:} Reconstruction quality for different model layers.
  Dashed lines are the performance bounds provided by ground-truth depth
  layers $(\bar{D}_1, \bar{D}_2, \bar{D}_3, \bar{D}_4, \bar{D}_5)$.
  \textit{Right:}  Accuracy of our model relative to the
  state-of-the-art, evaluated against objects only.  The upper and
  lower band indicate 75th and 25th quantiles. The higher variance of
  Tulsiani~\etal~\cite{tulsiani2018factoring} may be explained in part by
  the sensitivity of the model to having the correct initial set of object
  detections and pose estimates.
}
\label{fig:pr_objects}
\end{figure}

\vspace{-0.011in}
\subsection{Evaluation}
\vspace{-0.019in}
\label{sec:evaluation}

To reconstruct 3D surfaces from predicted multi-layer depth images as
well as the overhead height map, we first convert the depth images and height
maps into a point cloud and triangulate vertices that correspond to a
$2\times2$ neighborhood in image space. If the depth values of two adjacent
pixels is greater than a threshold $\delta \cdot a$, where $\delta$ is the
footprint of the pixel in camera coordinates and $a=7$, we do not create an edge between
those vertices. We do not predict 
the room envelope from the virtual overhead view, so only pixels with height
values higher than 5 cm above the floor are considered for reconstruction and
evaluation.

\boldhead{Metrics}
We use precision and recall of surface area as the metric to evaluate how
closely the predicted meshes align with the ground truth meshes, which is the native format for SUNCG and ScanNet.  Coverage is
determined as follows: We uniformly sample points on surface of the ground
truth mesh then compute the distance to the closest point on the predicted
mesh. We use sampling density $\rho = 10000/\text{meter}^2$ throughout our
experiments. Then we measure the percentage of inlier distances for given a
threshold.  \textit{Recall} is the coverage of the ground truth mesh by the
predicted mesh.  Conversely, \textit{precision} is the coverage of the
predicted mesh by the ground truth mesh.

\newcolumntype{Y}{>{\centering\arraybackslash\hsize=.5\hsize}X}
\begin{table}[t]
\centering
\begin{tabularx}{\columnwidth}{@{}XYY@{}}
 &     Precision & Recall\\ 
\hline
  $D_{1}$ &                    0.525    &                      0.212     \\ 
  $D_{1}$ \& Overhead &        \textbf{0.553}    & \textbf{0.275}     \\ \hline
  $D_{1,2,3,4}$ &                0.499    &                      0.417     \\ 
  $D_{1,2,3,4}$ \& Overhead &    \textbf{0.519}    & \textbf{0.457}     \\ \hline
\end{tabularx}
\vspace{-0.65em}
\caption{Augmenting the frontal depth prediction with the predicted virtual
  view height map improves both precision and recall (match threshold of 5cm).}
\label{fig:pr_table}
\end{table}

\boldhead{3D Scene Surface Reconstruction} To provide an upper-bound on the performance of our multi-layer depth
representation, we evaluate how well surfaces reconstructed from
ground-truth depths cover the full 3D mesh. This allows us to characterize the
benefits of adding additional layers to the representation.  Table
\ref{fig:ub_table} reports the coverage (recall) of the ground-truth at a
threshold of 5cm.  The left panels of Figure~\ref{fig:pr_objects} show a
breakdown of the precision and recall for the individual layers of our model
predictions along with the upper bounds achievable across a range of inlier
thresholds. 

Since the room envelope is a large component of many scenes, we also
analyze performance for objects (excluding the envelope).  Results 
summarized in Table \ref{fig:pr_table} show that the addition of
multiple depth layers significantly increases recall with only a small drop in
precision, and the addition of overhead EFT predictions boosts both precision and recall.

\boldhead{Ablation Study on Transformed Features}
To further demonstrate the value of the EFT module, we evaluate the accuracy
of the overhead height map prediction while incrementally excluding features.
We first exclude channels that correspond to the semantic segmentation network
features and compare the relative pixel-level L1 error.
We then exclude features from the depth prediction network,
using only RGB, frustum mask and best guess depth image.
This baseline corresponds to taking the prediction of the input view
model as an RGB-D image and re-rendering it from the virtual camera viewpoint.
The L1 error increases respectively from $0.132$ to $0.141$ and $0.144$,
which show that applying the EFT to the whole CNN feature map outperforms
simple geometric transfer.

\boldhead{Comparison to the State-of-the-art}
Finally, we compare the scene reconstruction performance of our end-to-end
approach with the object-based Factored3D \cite{tulsiani2018factoring}
method using their pre-trained weights, and converting voxel
outputs to surface meshes using marching cubes.  We evaluated on 3960 examples
from the SUNCG test set and compute precision and recall on objects surfaces
(excluding envelope).  As Figure~\ref{fig:pr_objects} shows, our method yields
roughly 3x improvement in recall and 2x increase in precision, providing
estimates which are both more complete and more accurate.  Figure~\ref{fig:soa}
highlights some qualitative differences.
To evaluate with an alternative metric, we voxelized scenes at 2.5cm$^3$
resolution (shown in Figure~\ref{fig:voxelization}). Using the voxel
intersection-over-union metric, we see significant performance improvements
over Tulsiani~\etal~\cite{drcTulsiani17} (0.102 to 0.310) on objects (see supplement for details).

\boldhead{Reconstruction on Real-world Images}
Our network model is trained entirely on synthetically generated
images~\cite{zhang2016physically}.  We test the ability of the model to
generalize to the NYUv2 dataset~\cite{SilbermanECCV12} via the promising 
comparison to Tulsiani~\etal\cite{tulsiani2018factoring} 
in Figure~\ref{fig:nyu}.

We additionally test our model on images from the ScanNetv2 dataset~\cite{dai2017scannet}. The dataset contains RGB-D image sequences taken in indoor scenes, as well as 3D reconstructions produced by
BundleFusion~\cite{dai2017bundlefusion}. For each video sequence from the 100 test scenes, we randomly sample $5\%$ of frames, and manually select 1000 RGB images to compare our algorithm to Tulsiani~\etal~\cite{tulsiani2018factoring}. 
We select images where the pose of the camera is almost perpendicular to the gravity orientation, the amount of motion blur is small, and the image does not depict a close-up view of a single object.
We treat the provided 3D reconstructions within each viewing frustum as ground truth annotations. 
As summarized in Table~\ref{fig:scannet_table}, our approach has significantly improved precision and recall to Tulsiani~\etal~\cite{tulsiani2018factoring}.

\vspace{-0.03in}
\section{Conclusion}
\vspace{-0.04in}
Our novel integration of deep learning and perspective geometry enables
complete 3D scene reconstruction from a single RGB image.  
We estimate multi-layer depth maps which model the front and back surfaces
of multiple objects as seen from the input camera, as well as the room envelope.
Our epipolar feature transformer network geometrically transfers input CNN 
features to estimate scene geometry from virtual viewpoints,
providing more complete coverage of real-world environments.
Experimental results on the SUNCG dataset~\cite{song2016ssc} demonstrate the 
effectiveness of our model. 
We also compare with prior work that predicts voxel representations of
scenes, and demonstrate the significant promise of our multi-view and multi-layer
depth representations for complete 3D scene reconstruction.

\boldhead{Acknowledgements}  
This research was supported by NSF grants IIS-1618806, IIS-1253538, CNS-1730158,
and a hardware donation from NVIDIA.

{\small
\bibliographystyle{ieee_fullname}
\bibliography{scene3d.bib}
}

\clearpage\mbox{}
\begin{center}
 {\large \textbf{Appendix}}
 \end{center}
\setcounter{section}{0}
\appendix

\section{System Overview}
We provide an overview of our 3D reconstruction system and additional
qualitative examples in our {\textbf{supplementary video}} (see project website).

\section{Training Data Generation}
\begin{figure}[!t]
\begin{center}
  \includegraphics[width=1\columnwidth]{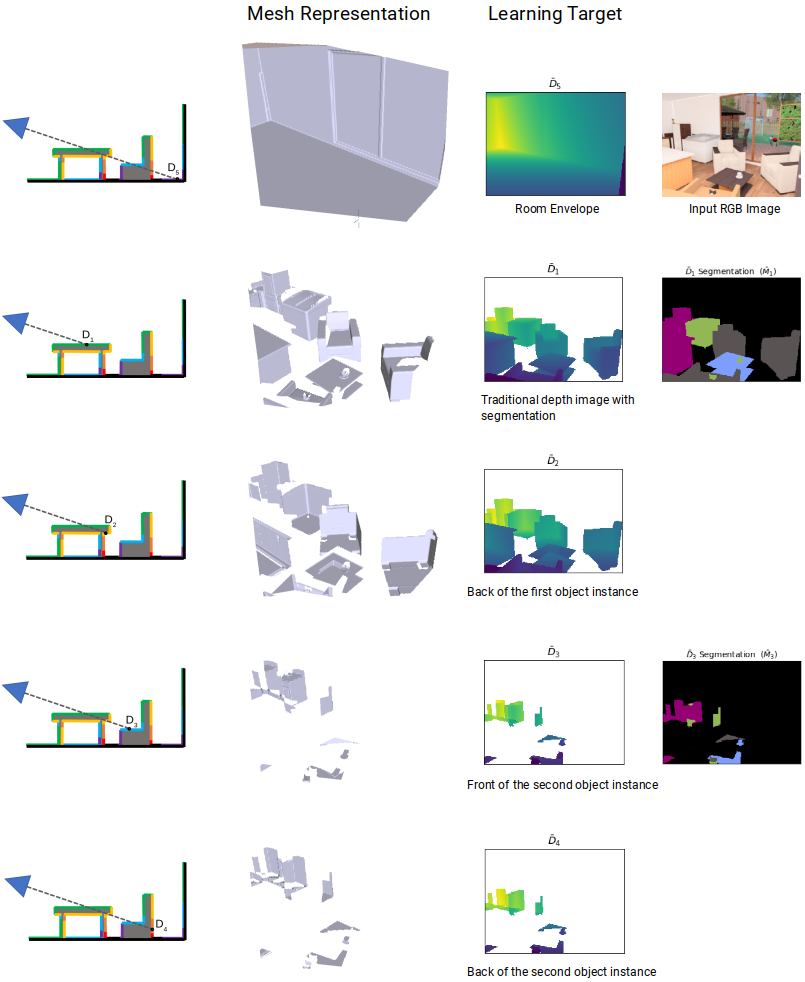}
\end{center}
\vspace{-0.15in}
    \caption{
	Layer-wise illustration of our multi-layer depth representation in
      3D. Table~\ref{fig:ub_table} in our paper reports an empirical analysis which shows that
      the five-layer representation ($\bar{D}_{1,2,3,4,5}$)
    covers 92\% of the scene geometry inside the viewing frustum.}
\label{fig:meshlayers}
\end{figure}

\begin{figure}[!t]
\begin{center}
\vspace{-0.03in}
  \includegraphics[width=1\columnwidth]{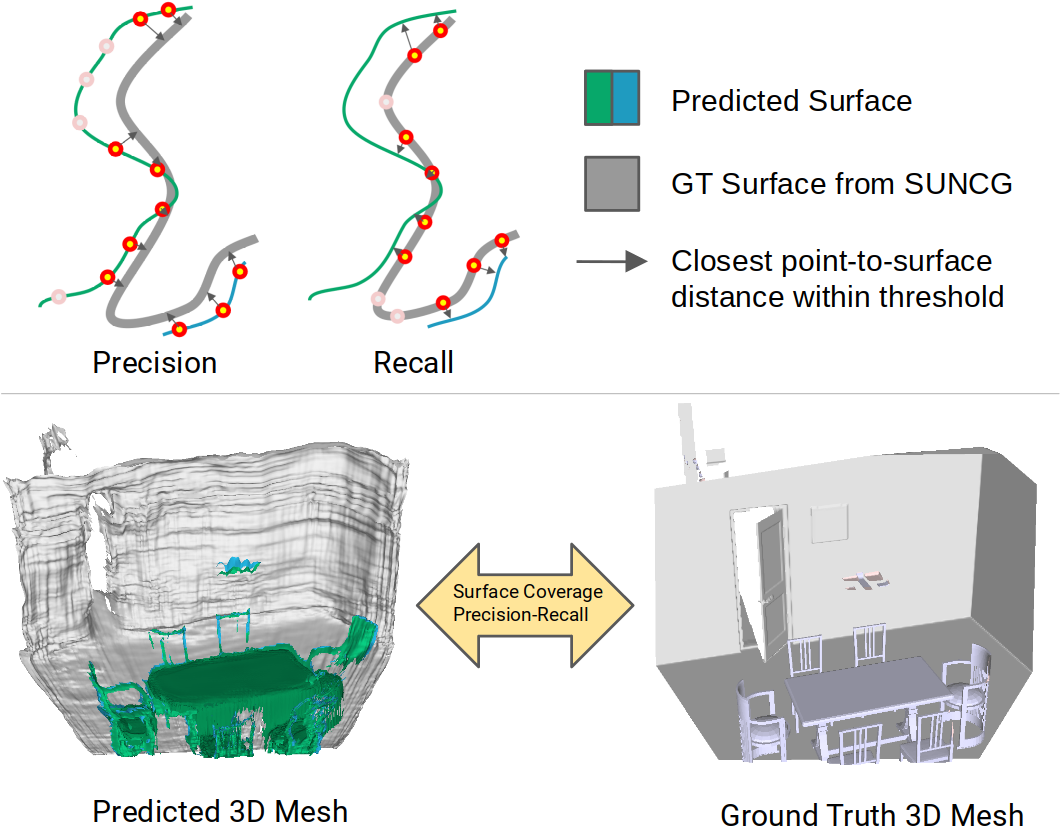}
\end{center}
\vspace{-0.1in}
    \caption{Illustration of our 3D precision-recall metrics.
    \textit{Top}: We perform a bidirectional surface coverage evaluation on the
    reconstructed triangle meshes. \textit{Bottom}: The ground truth mesh consists of
    all 3D surfaces within the viewing frustum and in front of the
    room envelope.
    We take the union of the predicted meshes from different views in world coordinates.
    This allows us to perform a layer-wise evaluation (e.g. Figure~\ref{fig:pr_objects} in our paper).
    }
\label{fig:evaluation}
\end{figure}

As we describe in Section~\ref{sec:trainingData} of our paper, we generate the target multi-layer depth
maps by performing multi-hit ray tracing on the ground-truth 3D mesh models.
If an object instance is completely occluded (i.e. not visible 
at all from the first-layer depth map), it is ignored in the subsequent layers.
The Physically-based Rendering~\cite{zhang2016physically} dataset ignores
objects in ``person'' and ``plant'' categories, so those categories are also
ignored when we generate our depth maps.
The complete list of room envelope categories (according to NYUv2 mapping)
are as follows: wall, floor, ceiling, door, floor\_mat, window, curtain, blinds,
picture, mirror, fireplace, roof, and whiteboard.
In our experiments, all room envelope categories are merged into a
single ``background'' category.
In Figure~\ref{fig:meshlayers}, we provide a layer-wise 3D visualization of our multi-layer depth representation.
Figure~\ref{fig:evaluation} illustrates our surface precision-recall metrics.

\begin{figure}[!ht]
\begin{center}
  \includegraphics[width=1\columnwidth]{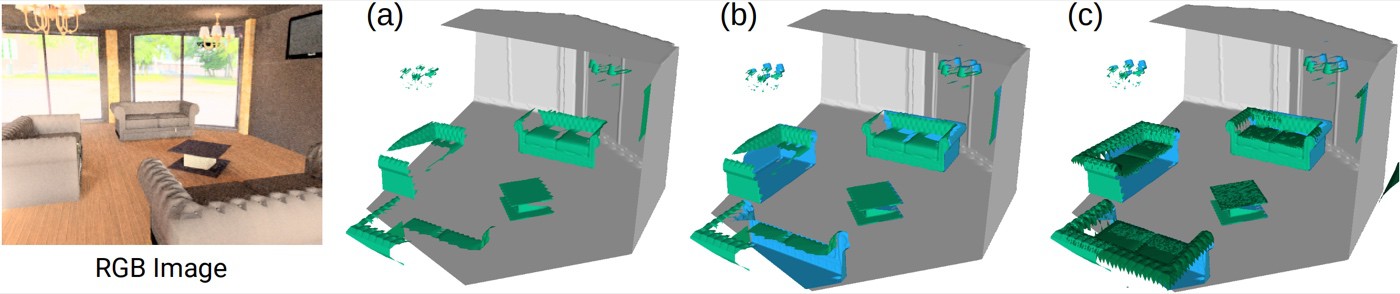}
\end{center}
\vspace{-0.12in}
    \caption{Illustration of ground-truth depth layers.
(a, b): 2.5D depth representation cannot accurately encode the
geometry of surfaces which are nearly tangent to the viewing direction.
(b): We model both the front and back surfaces of objects as seen from
the input camera.
(c): The tangent surfaces are sampled more densely in the additional virtual view (dark green).
Table~\ref{fig:pr_table} in our paper shows the effect of augmenting the frontal predictions with 
the virtual view predictions.
    }
\label{fig:gtlayers}
\end{figure}

\section{Representing the Back Surfaces of Objects}
Without the back surfaces, ground truth depth layers
($\bar{D}_{1,3,5}$) cover only 82\% of the scene geometry inside the viewing
frustum (vs. 92\% including frontal surfaces --- refer to Table~\ref{fig:ub_table} in our paper for full comparison).
Figure~\ref{fig:gtlayers}(a) visualizes $\bar{D}_{1,3,5}$, without the back surfaces.
This representation, \textit{layered depth image} (LDI)~\cite{shade1998layered}, was originally
developed in the computer graphics community~\cite{shade1998layered}
as an algorithm for rendering textured depth images using parallax
transformation.
Works based on prediction of LDI or its variants \cite{tulsiani2018layer, zhou2018stereo}
therefore do not represent the invisible back surfaces of objects.
Prediction of back surfaces enables volumetric inference in our epipolar transformation.

\section{Multi-layer Depth Prediction}
\begin{figure*}[!t]
\centering
\includegraphics[page=1, trim={6cm 32cm 19cm 8cm},clip, scale=0.335]{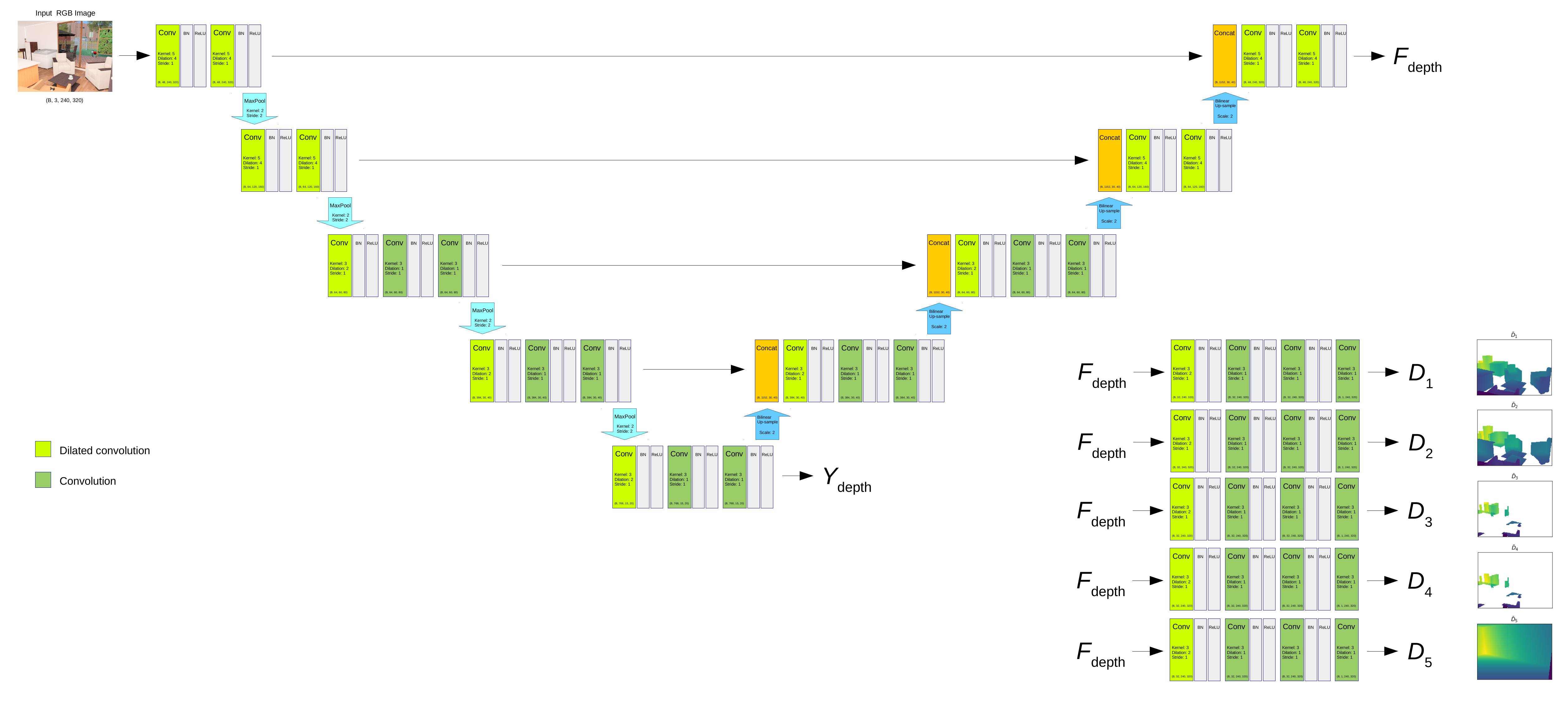}
\caption{Network architecture for multi-layer depth prediction. The horizontal
arrows in the network represent skip connections. This figure, along with
following figures, is best viewed in color and on screen.}
\label{fig:network1}
\end{figure*}

See Figure~\ref{fig:network1} for network parameters of our multi-layer depth
prediction model.  All batch normalization layers have momentum 0.005, and all
activation layers are Leaky ReLUs layers with $\alpha=0.01$.  We use In-place
Activated BatchNorm~\cite{rotabulo2017place} for all of our batch normalization
layers. We trained the network for 40 epochs.

\section{Multi-layer Semantic Segmentation}
\label{sec:mls}
\begin{figure*}[!ht]
\centering
\includegraphics[page=2, trim={6cm 38.5cm 19cm 8cm},clip, scale=0.335]{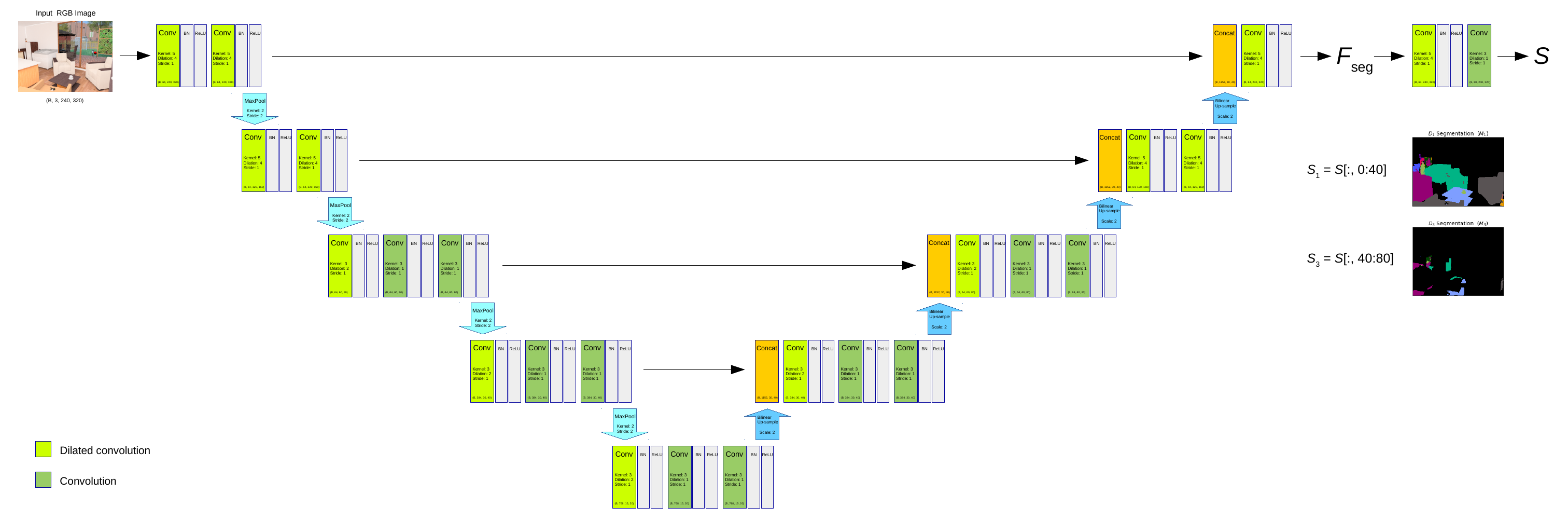}
\caption{Network architecture for multi-layer semantic segmentation network.
(Best viewed in color and on screen)}
\label{fig:network2}
\end{figure*}

See Figure~\ref{fig:network2} for network parameters of multi-layer semantic
segmentation.  We construct a binary mask for all foreground objects, and
define segmentation mask $M_l$ as all non-background pixels at layer $l$.  As
mentioned in section~3.1, $D_1$ and $D_2$ have the same segmentation due to
symmetry, so we only segment layers $1$ and $3$.  The purpose of the foreground
object labels is to be used as a supervisory signal for feature extraction
$F_{\text{seg}}$, which is used as input to our Epipolar Feature Transformer
Networks.

\section{Virtual-view Prediction}
\begin{figure*}[!ht]
\centering
\includegraphics[page=4, trim={5.5cm 38.5cm 20cm 8cm},clip, scale=0.335]{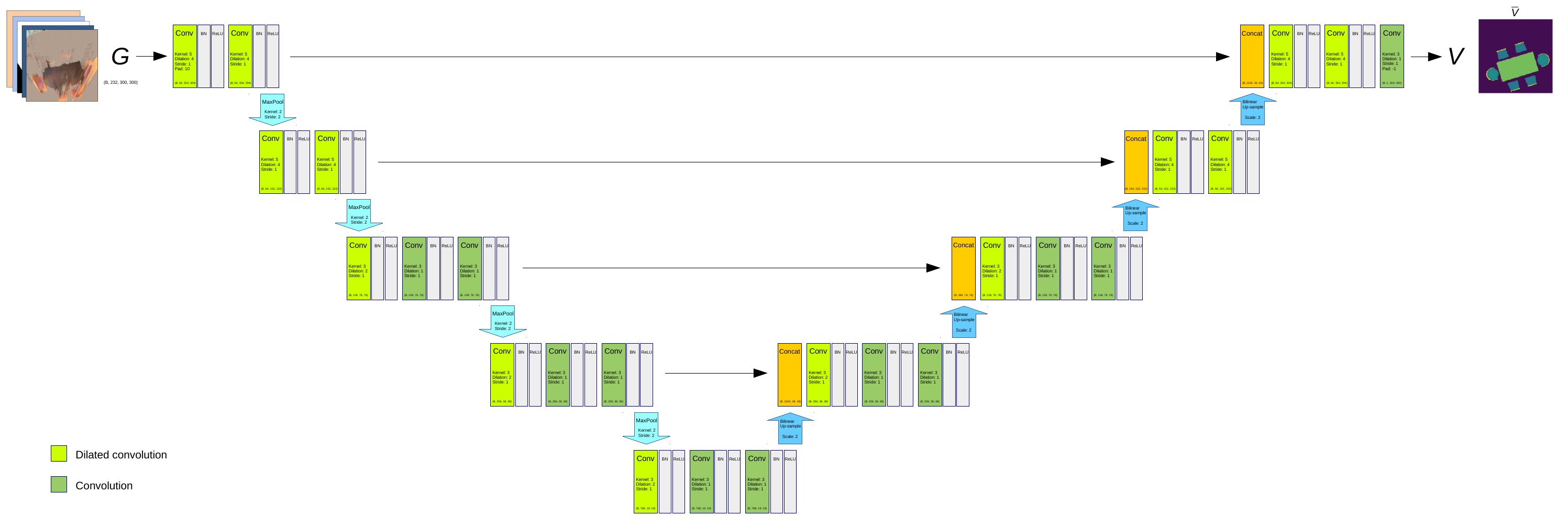}
\caption{Network architecture for virtual view surface prediction network. (Best viewed
in color and on screen) }
\label{fig:network4}
\end{figure*}

\begin{figure*}[!ht]
\centering
\includegraphics[page=5, trim={5.5cm 38.5cm 20cm 8cm},clip, scale=0.335]{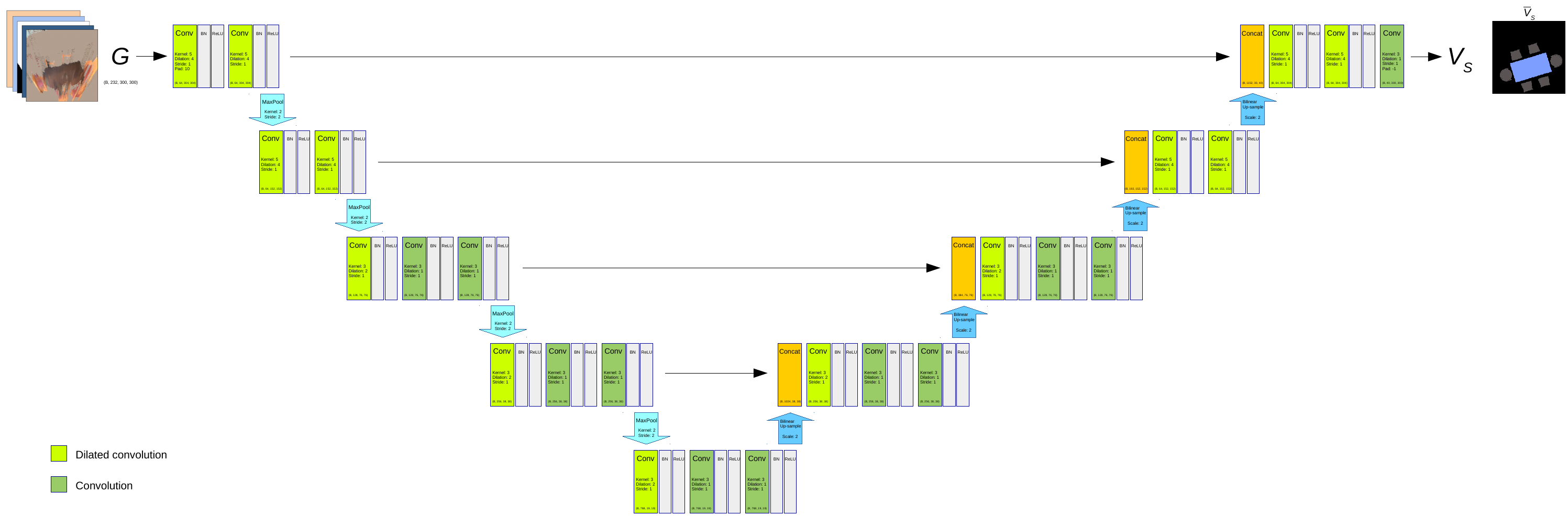}
\caption{Network architecture for virtual view segmentation network. (Best viewed
in color and on screen) }
\label{fig:network5}
\end{figure*}

See Figure~\ref{fig:network4} and \ref{fig:network5} for network parameters of
our virtual-view height map prediction and segmentation models.
The height map prediction network is trained to minimize foreground pixel losses.
At test time, the background mask predicted by the segmentation network is used
to zero out the floor pixels. The floor height is assumed to be zero in world
coordinates. An alternate approach is minimizing both foreground and background
losses and thus allowing the height map predictor to implicitly segment the floor by
predicting zeros.  We experimented with both architectures and found the
explicit segmentation approach to perform better.

\begin{figure}
	\centering
	\begin{tabular}{ccc}
\subfigimg[width=0.30\columnwidth]{ScanNet}{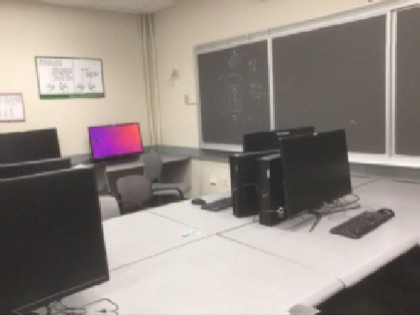}&
\subfigimg[width=0.26\columnwidth]{[Tulsiani, 2018]}{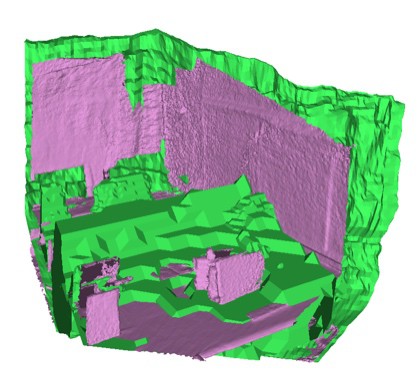}&
\subfigimg[width=0.28\columnwidth]{Ours}{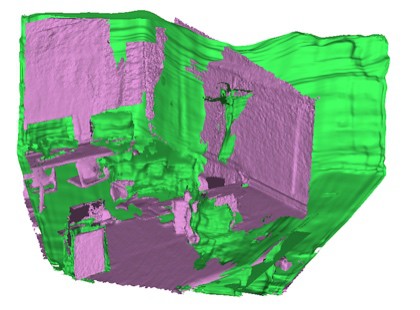}\\
\subfigimg[width=0.30\columnwidth]{ScanNet}{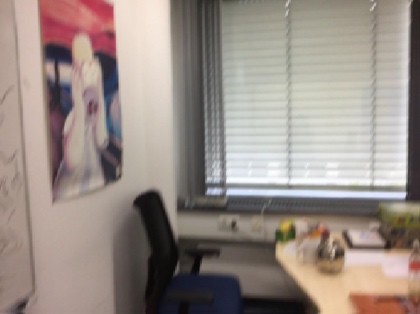}&
\subfigimg[width=0.22\columnwidth]{[Tulsiani, 2018]}{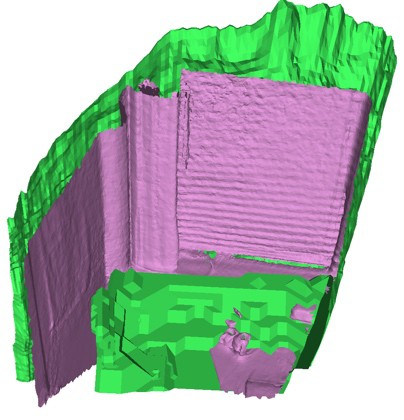}&
\subfigimg[width=0.25\columnwidth]{Ours}{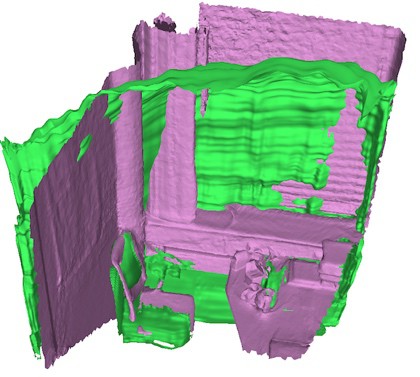}\\
\end{tabular}
\caption{Evaluation of 3D reconstruction on the ScanNet~\cite{dai2017scannet} dataset, where green regions are detected
objects and pink regions are ground truth.}
\label{fig:scannetextra}
\end{figure}

\begin{figure}
\begin{center}
  \includegraphics[width=1\columnwidth]{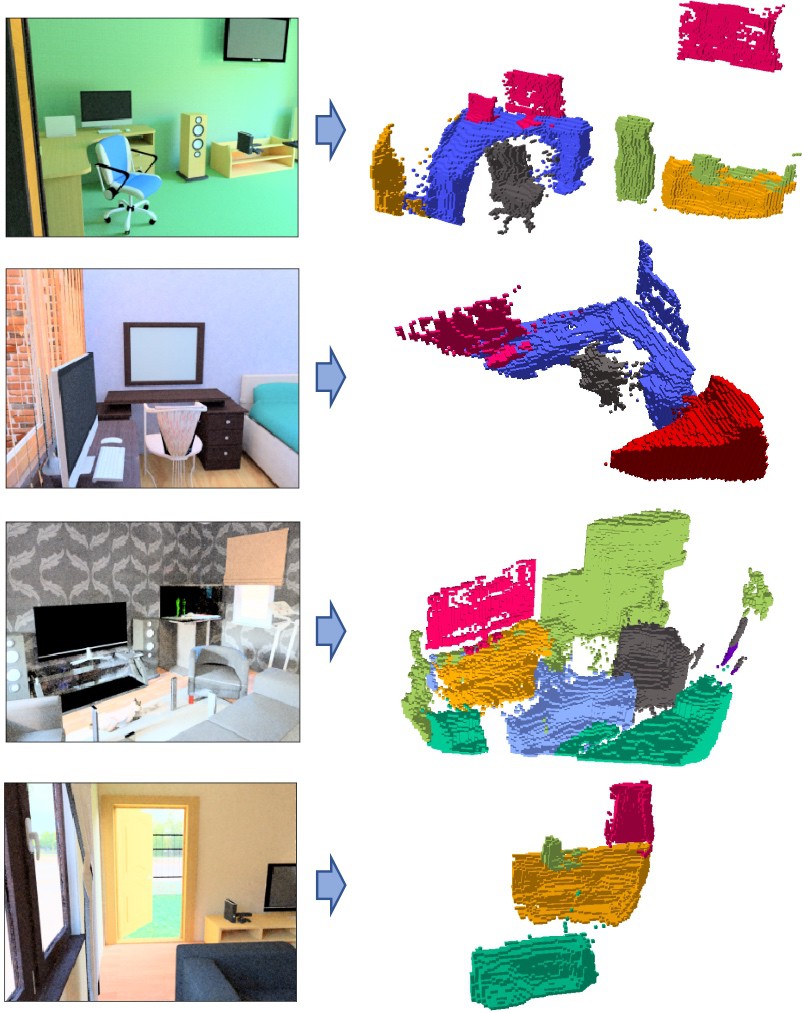}
\end{center}
\vspace{-0.2in}
    \caption{Volumetric evaluation of our predicted multi-layer depth maps on the SUNCG~\cite{song2016ssc} dataset.}
\label{fig:extravoxels}
\end{figure}

\begin{figure}
\begin{center}
\vspace{-0.03in}
  \includegraphics[width=1\columnwidth]{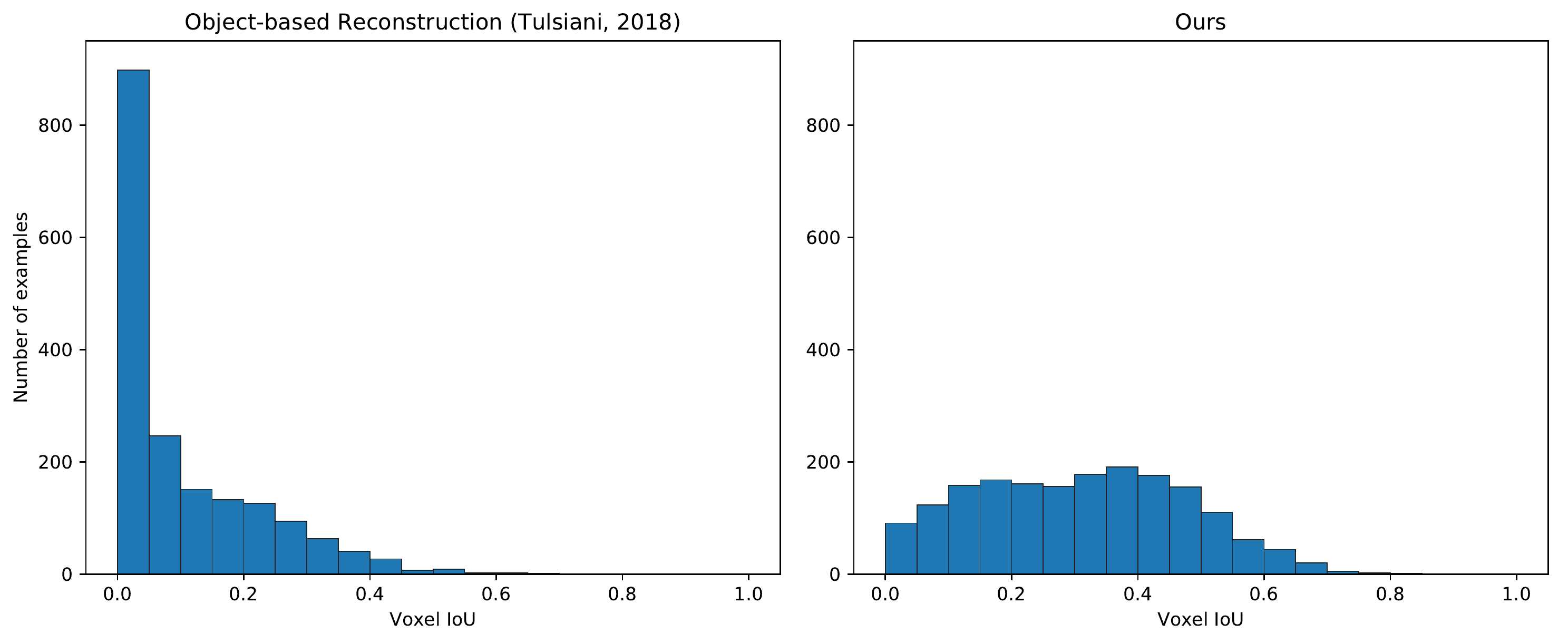}
\end{center}
\vspace{-0.13in}
    \caption{Distribution of voxel intersection-over-union on SUNCG (excluding room layouts).
	We observe that object-based reconstruction is sensitive to detection failure and misalignment on thin structures.}
\label{fig:voxelhist}
\end{figure}

\section{Voxelization of Multi-layer Depth Prediction}
Given a 10m$^3$ voxel grid of resolution 400 (equivalently, 2.5cm$^3$) with a bounding box ranging from (-5,-5,-10) to (5,5,0) in camera space, we project the center of each voxel into the predicted depth maps. If the depth value for the projected voxel falls in the first object
interval $({D}_1, {D}_2)$ or the occluded object interval $({D}_3, {D}_4)$, the
voxel is marked occupied. We evaluate our voxelization against the SUNCG ground
truth object meshes inside the viewing frustum, voxelized using the Binvox software
which implements z-buffer based carving and parity voting methods. 
We also voxelize the predicted Factored3D \cite{tulsiani2018factoring} objects
(same meshes evaluated in Figure~\ref{fig:pr_objects} of our paper) using Binvox under the same setting as the ground truth.
We randomly select 1800 examples from the test set and compute the intersection-over-union of all objects in the scene.
In addition to Figure~\ref{fig:voxelization} of our paper, Figure~\ref{fig:extravoxels} shows 
a visualization of our voxels, colored according to the predicted semantic labeling.
Figure~\ref{fig:voxelhist} shows a histogram of voxel intersection-over-union values.

\section{Predictions on NYU and SUNCG}
\begin{figure}
\centering
\begin{tabular}{ccc}
\subfigimg[width=0.30\columnwidth]{Input}{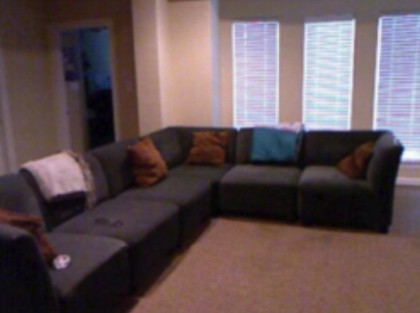}&
\subfigimg[width=0.30\columnwidth]{[Tulsiani, 2018]}{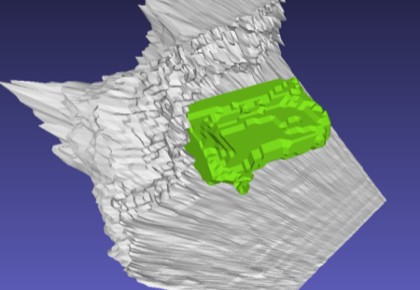}&
\subfigimg[width=0.30\columnwidth]{Ours}{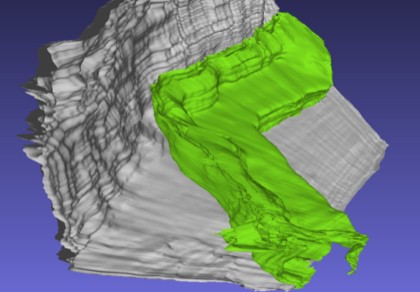}\\
\subfigimg[width=0.30\columnwidth]{Input}{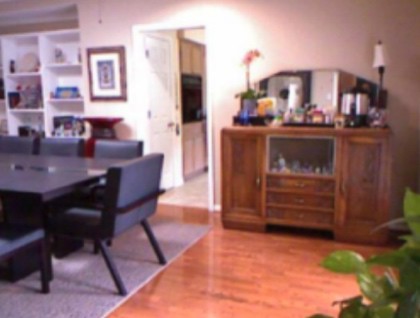}&
\subfigimg[width=0.30\columnwidth]{[Tulsiani, 2018]}{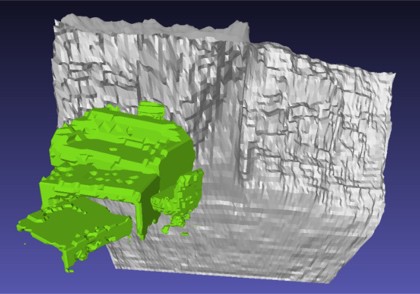}&
\subfigimg[width=0.30\columnwidth]{Ours}{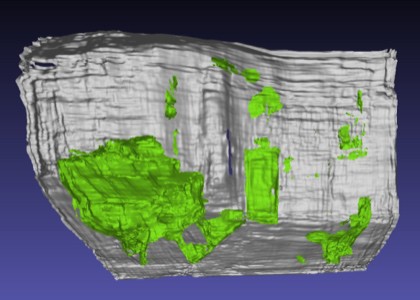}\\
\subfigimg[width=0.30\columnwidth]{Input}{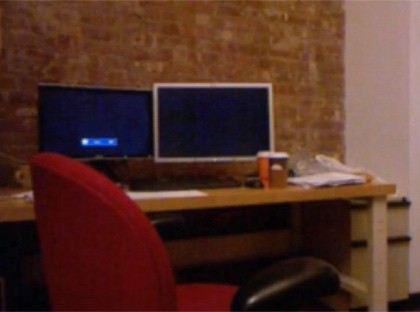}&
\subfigimg[width=0.30\columnwidth]{[Tulsiani, 2018]}{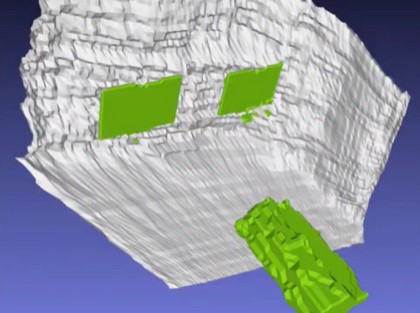}&
\subfigimg[width=0.30\columnwidth]{Ours}{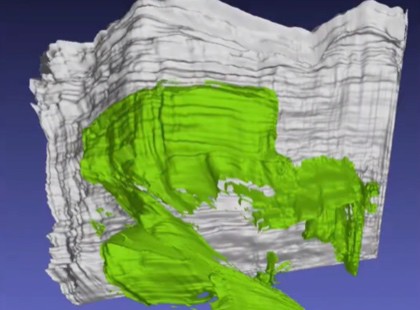}\\
\subfigimg[width=0.30\columnwidth]{Input}{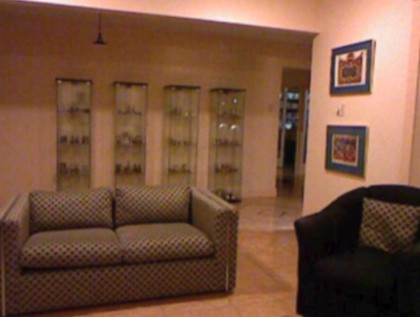}&
\subfigimg[width=0.30\columnwidth]{[Tulsiani, 2018]}{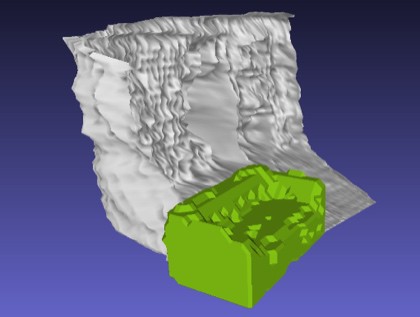}&
\subfigimg[width=0.30\columnwidth]{Ours}{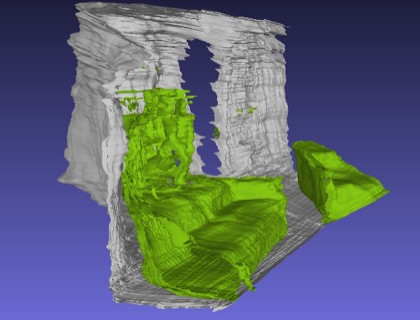}\\
\subfigimg[width=0.30\columnwidth]{Input}{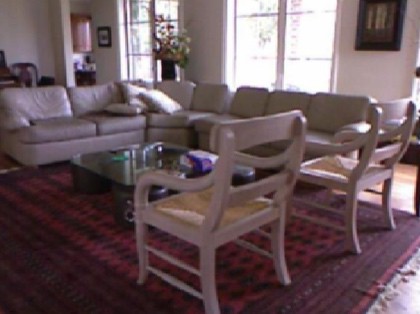}&
\subfigimg[width=0.30\columnwidth]{[Tulsiani, 2018]}{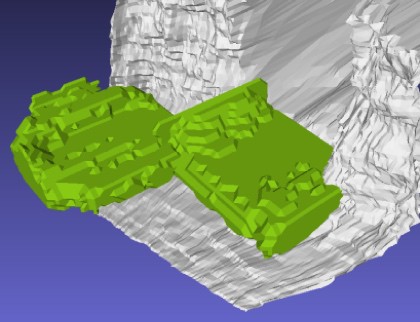}&
\subfigimg[width=0.30\columnwidth]{Ours}{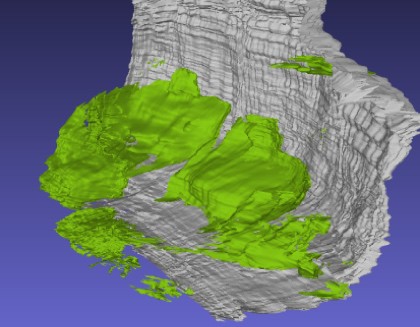}\\
\end{tabular}
\caption{Evaluation of 3D scene reconstruction on the NYUv2~\cite{SilbermanECCV12} dataset.}
\label{fig:nyu2}
\end{figure}

\begin{figure}
\centering
\begin{tabular}{ccc}
\subfigimg[width=0.30\columnwidth]{Input}{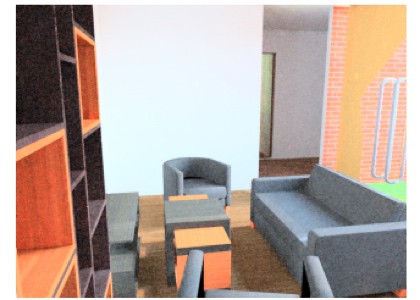}&
\subfigimg[width=0.30\columnwidth]{3D Reconstruction}{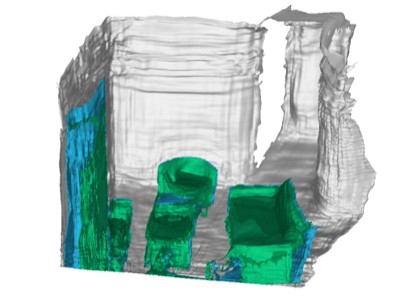}&
\subfigimg[width=0.30\columnwidth]{3D Reconstruction}{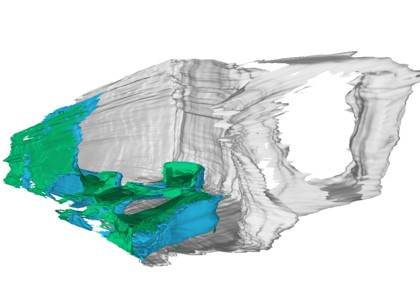}\\
\subfigimg[width=0.30\columnwidth]{Input}{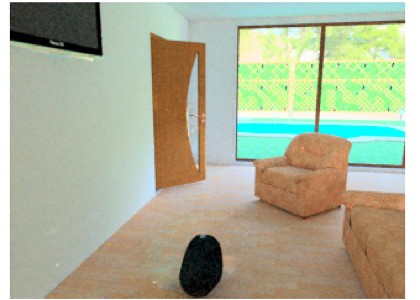}&
\subfigimg[width=0.30\columnwidth]{3D Reconstruction}{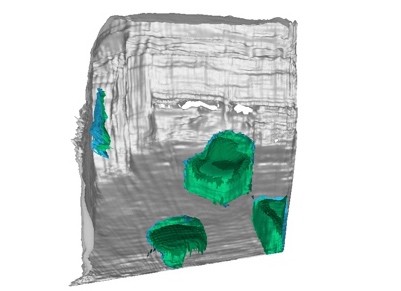}&
\subfigimg[width=0.30\columnwidth]{3D Reconstruction}{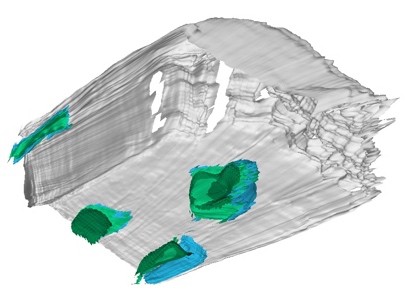} \\
\subfigimg[width=0.30\columnwidth]{Input}{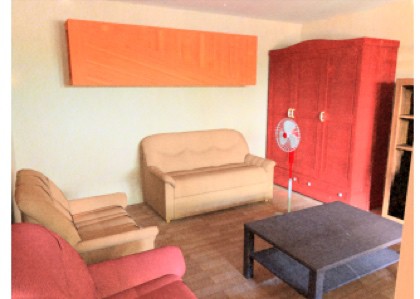}&
\subfigimg[width=0.30\columnwidth]{3D Reconstruction}{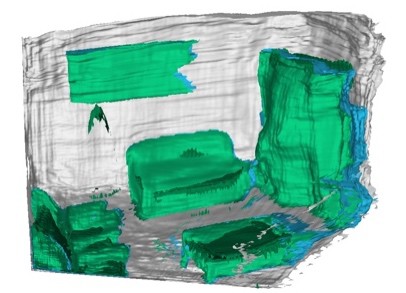}&
\subfigimg[width=0.30\columnwidth]{3D Reconstruction}{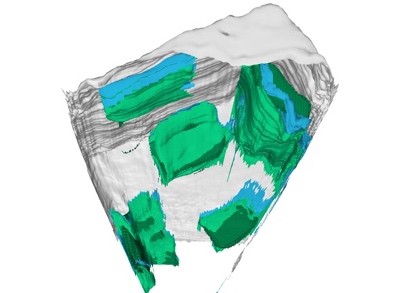} \\
\subfigimg[width=0.30\columnwidth]{Input}{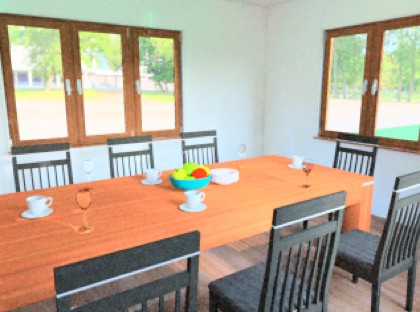}&
\subfigimg[width=0.30\columnwidth]{3D Reconstruction}{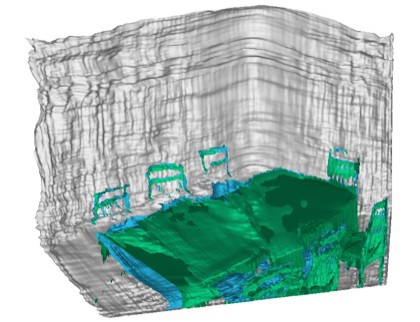}&
\subfigimg[width=0.30\columnwidth]{3D Reconstruction}{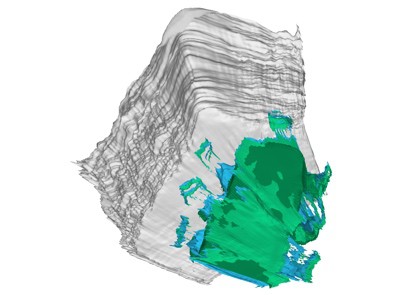} \\
\end{tabular}
\caption{Evaluation of 3D scene reconstruction on the SUNCG~\cite{song2016ssc} dataset.}
\label{fig:exp_recon2}
\end{figure}

Figures~\ref{fig:exp_recon2} and \ref{fig:nyu2} show additional 3D scene reconstruction results.
We provide more visualizations of our network outputs and error maps on the SUNCG dataset in
the last few pages of the supplementary material.

\onecolumn

\begin{figure*}[!ht]
\centering
\includegraphics[page=3, trim={6cm 48cm 42cm 8cm},clip, scale=0.40]{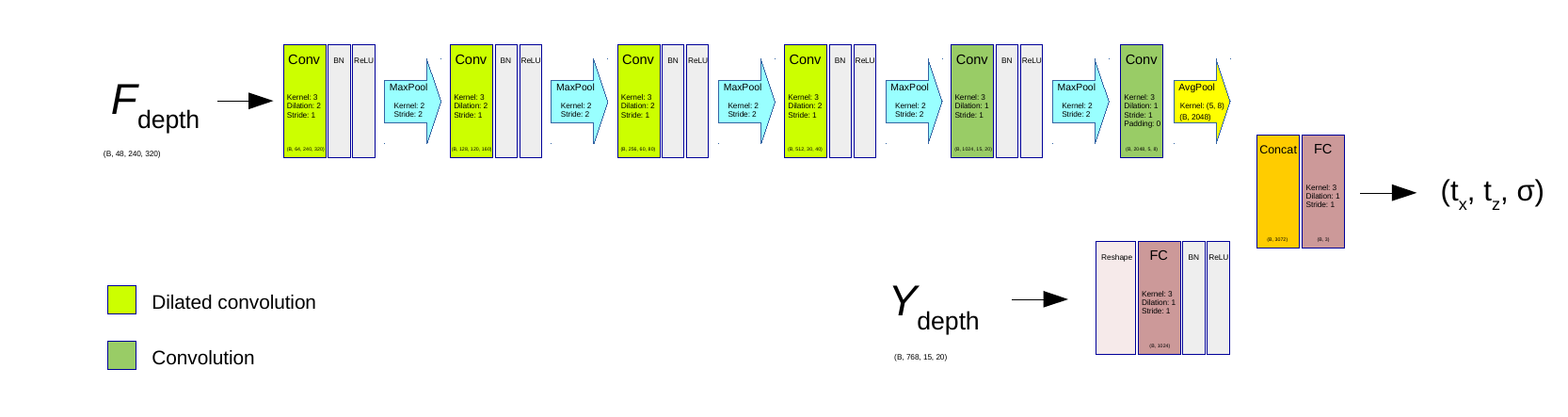}
\caption{ Network architecture for virtual camera pose proposal network. (Best
viewed in color and on screen) }
\label{fig:network3}
\end{figure*}

\clearpage
{
\includegraphics[page=5, clip, scale=0.55]{figures/depth-qualitative-lowres.pdf}

\includegraphics[page=6, clip, scale=0.55]{figures/depth-qualitative-lowres.pdf}

\includegraphics[page=2, clip, scale=0.55]{figures/depth-qualitative-lowres.pdf}

\includegraphics[page=12, clip, scale=0.55]{figures/depth-qualitative-lowres.pdf}

\includegraphics[page=3, clip, scale=0.55]{figures/depth-qualitative-lowres.pdf}

\includegraphics[page=1, clip, scale=0.55]{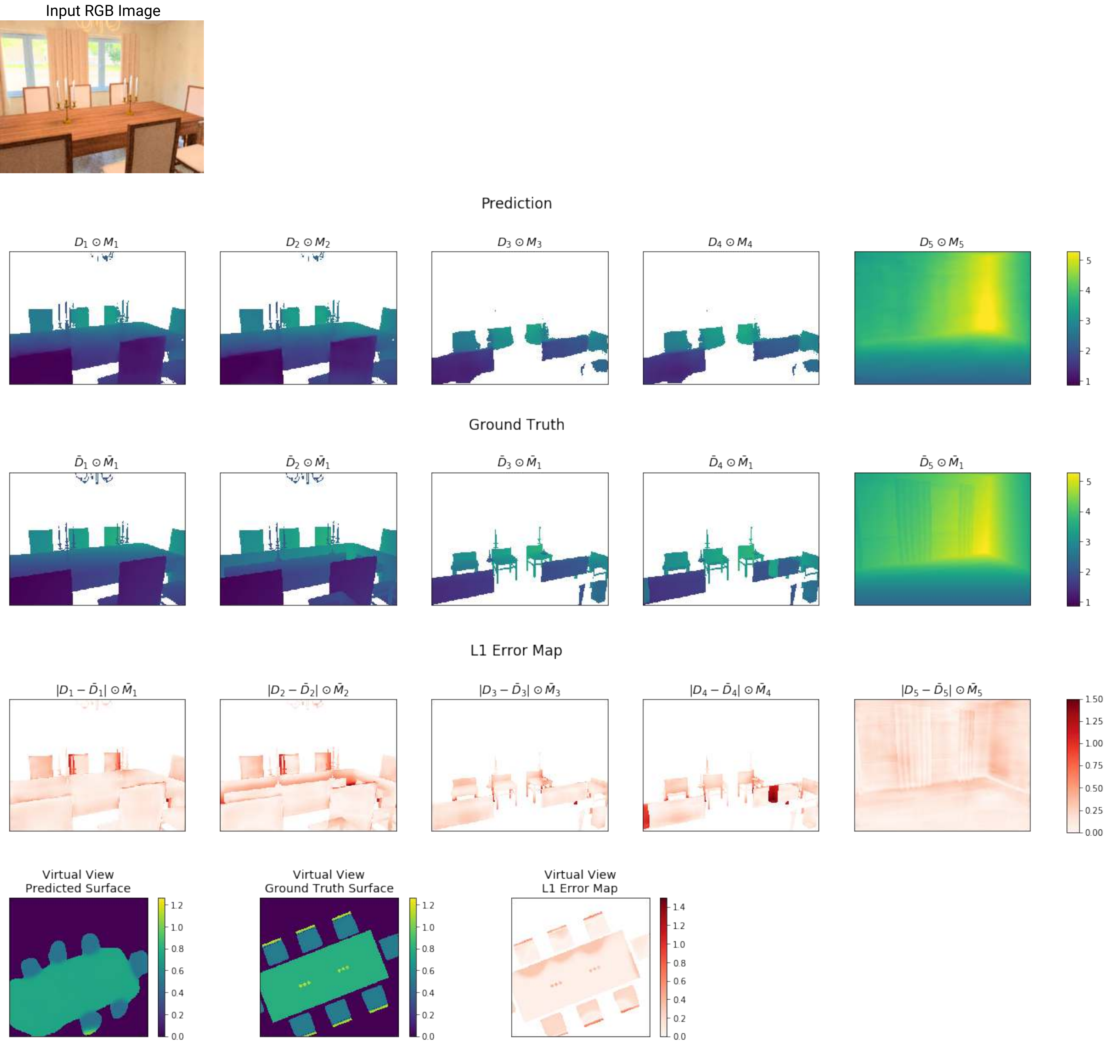}

\includegraphics[page=4, clip, scale=0.55]{figures/depth-qualitative-lowres.pdf}

\includegraphics[page=7, clip, scale=0.55]{figures/depth-qualitative-lowres.pdf}

\includegraphics[page=8, clip, scale=0.55]{figures/depth-qualitative-lowres.pdf}

\includegraphics[page=9, clip, scale=0.55]{figures/depth-qualitative-lowres.pdf}

\includegraphics[page=10, clip, scale=0.55]{figures/depth-qualitative-lowres.pdf}

\includegraphics[page=11, clip, scale=0.55]{figures/depth-qualitative-lowres.pdf}
}

\end{document}